\documentclass{article}

\usepackage[preprint,nonatbib]{neurips_2023}

\usepackage[utf8]{inputenc} 
\usepackage[T1]{fontenc}    
\usepackage{hyperref}       
\usepackage{url}            
\usepackage{booktabs}       
\usepackage{amsfonts}       
\usepackage{nicefrac}       
\usepackage{microtype}      
\usepackage{xcolor}         
\usepackage{graphicx}

\usepackage{lscape}

\title{Analysis of Atom-level pretraining with QM data for 
Graph Neural Networks Molecular property  
models}

\author{%
 Jose Arjona-Medina \\
  Janssen Research \& Development\\
  \texttt{jarjonam@its.jnj.com} \\
   \And
  Ramil Nugmanov \\
   Janssen Pharmaceutica NV\\
   \texttt{rnugmano@its.jnj.com} \\
}

\begin{document}

\maketitle

\begin{abstract}
Despite the rapid and significant advancements in deep learning for 
Quantitative Structure-Activity Relationship (QSAR) models, 
the challenge of learning robust molecular representations 
that effectively generalize in real-world scenarios to novel compounds 
remains an elusive and unresolved task.  
This study examines how atom-level pretraining with quantum mechanics (QM) data 
can mitigate violations of assumptions regarding the distributional similarity 
between training and test data and therefore improve performance and generalization 
in downstream tasks. 
In the public dataset Therapeutics Data Commons (TDC), we show how pretraining on atom-level QM 
improves performance overall and makes the activation of the features distributes more Gaussian-like which results in a representation that is more robust to distribution shifts. 
To the best of our knowledge, this is the first time that 
hidden state molecular representations are analyzed 
to compare the effects of molecule-level and atom-level pretraining on QM data.  

\end{abstract}

\section{Introduction}

Despite the rapid and significant advancements in deep learning for Quantitative Structure-Activity Relationship (QSAR) models, 
the challenge of learning robust molecular representations that effectively generalize in real-world scenarios to novel compounds remains an elusive and unresolved task.

At the core of supervised learning with deep neural networks, 
there is the underlying assumption that the function we aim to approximate is "smooth" enough, 
meaning that small changes in the input should not result in large changes in the output \cite{Goodfellow-et-al-2016}.
However, in the application field of QSAR, 
we often experience regions where small chemical modifications 
drastically change the biological response \cite{Cruz-Monteagudo:14}.
For example, the extreme forms of Structure-Activity Relationship (SAR) 
discontinuity are called activity cliffs (AC)  \cite{Bajorath:09}, 
which are formed by pairs of structurally similar compounds 
with large differences in potency  [\cite{Stumpfe:12},\cite{Peltason:11}].
One way of overcoming this problem would be to increase the number of datapoints 
since the smoothness assumption works well as long as there are enough examples 
for the learning algorithm to observe high points on most peaks and low points 
on most valleys of the true underlying function to be learned \cite{Goodfellow-et-al-2016}.
However, it is not a general or scalable solution 
because the costs of producing additional experimental datapoints are high 
and would be applicable only locally for a specific target property
%

Supervised Learning methods also assume that 
test data comes from the same underlying distribution as training data \cite{Bishop:06}. 
Although novel techniques often present impressive metrics, 
most often they do not suffice to meet the practical needs in real-world drug discovery \cite{Deng:23}.
In the real world, data production is by construction strongly biased: 
Chemists work with structurally analogous series. 
In practice, target data is drifting \cite{Tossou:23},
making these models obsolete for real-world production environments.
The fundamental factors that influence molecular properties are still unexplored \cite{Deng:23}.
With statistical analyses, fixed representations like fingerprints 
generally match end-to-end deep learning (DL) models in most datasets \cite{Deng:23, Baptista:22,Robinson:20},
especially in the presence of activity cliffs \cite{vanTilborg:22}.
Much of the progress in molecular property estimation 
has been driven by the strategic incorporation of inductive biases \cite{Li:20, Xia:23}. 
Techniques such as pretraining \cite{Xia:23b}, the use of equivariant networks, 
and graph neural networks have proven useful \cite{Satorras:21,Le:22}. 
The shift towards leveraging domain-specific knowledge through inductive biases 
underlines the evolution in our approach to QSAR modeling.


\paragraph{Pretraining and Transfer Learning in QSAR Models.} 
Pretraining has emerged as a potent solution 
to overcome data scarcity and enhance model generalizability in various domains, 
including image and natural language processing. 
However, without appropriate domain expertise, 
pretraining can sometimes lead to negative transfer 
or only marginal performance gains \cite{Rosenstein:05}. 
In the realm of molecular modeling, pretraining strategies have been successful, 
particularly when models are pretrained on large-scale datasets \cite{Wang:19,Chithrananda:20,Stark:23,Xu:24}.
For instance, the large-scale pretraining of models has been shown to improve sample efficiency 
in active learning frameworks used for molecule virtual screening \cite{Cao:23}.
Or \cite{Schimunek:23} which enriches molecule representations with contextual knowledge from reference molecules
using a Modern Hopfield Network \cite{Ramsauer:21}.
Combining node-level and graph-level pre-training tasks to achieve superior generalization, 
particularly in out-of-distribution scenarios was presented in \cite{Hu:20}.
ChemBERTa utilizes large-scale pretraining datasets 
to explore the effects of dataset size on downstream task performance \cite{Chithrananda:20}. 

\paragraph{Multitask learning.} Closely related to transfer learning, 
involves the simultaneous learning of multiple related tasks, 
enhancing the model's ability to generalize across tasks \cite{Caruana:97}. 
This approach has been leveraged effectively in models like MolPMoFiT, 
which was pretrained using bioactive molecules from ChEMBL for QSPR/QSAR tasks \cite{Li:20}.

\paragraph{Multimodal and Self-supervised Learning.}
Recent advancements have also been made in multimodal learning and self-supervised learning frameworks. 
The GIMLET model, for example, pretrained on a dataset of molecule tasks with textual instructions, 
bridging the gap between graph and text data modalities 
and paving the way for instruction-based pretraining \cite{Zhao:23}. 
Similarly, the D\&D framework utilizes a cross-modal distillation approach, 
transferring knowledge from 3D to 2D molecular structures, 
thereby significantly enhancing model performance in downstream tasks \cite{Cho:23}.
Or denoising 3D structures as a pre-training objective, 
setting new benchmarks in the widely used QM9 dataset \cite{Zaidi:22}.

\paragraph{Contrastive learning} approaches \cite{Le-Khac:20,Khosla:20} 
can overcome data limitations by integrating additional data.
For example, CLOOME \cite{Sanchez-Fernandez:23}
embeds bioimages and chemical structures into a unified space, 
enabling highly accurate retrieval of bioimages based on chemical structures.
MolFeSCue \cite{Zhang:24} incorporates a dynamic contrastive loss function tailored for class imbalance.
MolCLR \cite{Wang:22} uses large-scale unlabeled data, graph-based augmentations, and a contrastive learning strategy to improve model generalization.
A recent overview of methods for multi-modal contrastive learning
is presented in \cite{Seidl:24}.

Our approach effectively combines pretraining with QM data at node-level,
to improve the downstream task of property modeling.
We analyze the effect of this pretraining in the distribution of the features of the network,
and we observe that atom-level pretrained networks have more Gaussian-like distributions
which are more robust to distribution shifts in the input space,
establishing a direct connection between pretraining at node-level and 
supervised learning theory of generalization.

Or contributions can be summarized as follows:
\begin{itemize}
\item We empirically show that atom-level pretraining in Graph neural networks with QM properties improves performance over scratch networks and molecular-level pretraining, in the TDC public dataset. 
\item We empirically show that atom-level pretraining produces a more normal distribution of features, compared to scratch and molecular-level pretrained networks.
\item We empirically show that atom-level pretraining produces a more robust molecular representation against distribution shift from train to eval and train to test datasplits, which could explain the performance gain of atom-level pretraining approach.
\end{itemize}
To the best of our knowledge, this is the first study that aims to understand 
how atom-level pretraining improves molecular representation.
%

\section{Methods}

\paragraph{Graphormer.}
In 2021, Microsoft published Graphormer\cite{Ying:21}, a neural network specifically engineered for processing graphs and, more pointedly, molecular structures. This network has exhibited outstanding performance in molecular quantum properties, adapting the architecture of BERT\cite{Devlin:19} to suit graph data effectively.
Graphormer introduces several novel features, the key among them being "centrality encoding". This technique captures the importance of nodes within the graph by integrating graph degree centrality, which is encoded as embedding vectors added into the atom's node type embeddings.
Additionally, Graphormer incorporates a "spatial encoding" strategy. This involves representing the shortest path between pairs of nodes, which is then utilized as a learnable bias in the attention matrix. This concept mirrors techniques used in other advanced models like T5\cite{Raffel:19} and ALIBI\cite{Press:21}, highlighting its relevance and utility in enhancing the model's attention mechanism within graph networks.

\paragraph{Custom implementation.}
Following \cite{Nugmanov:22}, we have refined the "centrality encoder" of the original Graphormer to encompass not just explicit neighbors but total neighbors, integrating both explicit atom neighbors and implicit hydrogens. This approach eliminates the need for an "edge encoder", leveraging a combination of the centrality encoder and an atom type encoder to implicitly capture atom hybridization. Moreover, we have streamlined the model by omitting the encoding of atoms' formal charges and radical states, allowing for the representation of resonance structures in a unified form and dispensing with the traditional concept of "aromatic" bonds typically applied to arenes. Additionally, our model introduces a "spatial encoder" constrained by a configurable maximum distance threshold to enhance computational efficiency and model accuracy. By treating distances beyond this threshold uniformly, the model focuses on more relevant short-range interactions understanding molecular structures.
\begin{figure}
    \centering
    \includegraphics[width=0.5\linewidth]{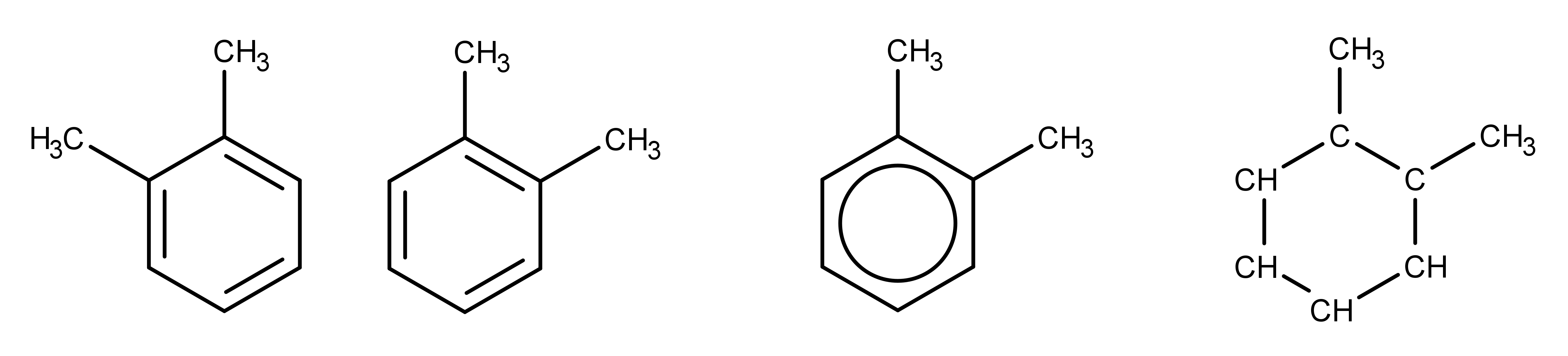}
    \caption{Different aromatic ring representations folded into a single form}
    \label{fig:enter-label}
\end{figure}

\paragraph{Enabling multitask learning by task-specific virtual nodes.}
Similar to the original BERT model, Graphormer utilizes a [CLS] token as a global readout for estimating properties of graphs and molecules. This methodology aligns with techniques used in graph convolutional networks (GCNs), where virtual nodes connected to every node of the graph can serve a comparable purpose \cite{Cai:23}. Building on this concept, we propose a novel extension tailored for multitask learning applications. In our approach, we employ the same encoder and output head to estimate multiple properties, but differentiate the tasks by using distinct virtual nodes for each. This allows for task-specific processing while maintaining a unified architecture, enhancing the model's efficiency, adaptability, and scalability in handling diverse learning tasks.


\paragraph{\textbf{Quantitative Assessment of learned features.}}
To analyze the impact of pretraining methods
on the molecular representation captured by the model,
we conducted a comprehensive analysis of learned feature distributions.
We run the models on the three different data splits (training, validation, and test),
and extract the distribution of features 
after the first layer of the Graphormer architecture.

For each compound in the dataset, 
we aggregated the node representations after the first Graphormer layer, 
resulting in a cumulative dataset-wide node representation. 
This aggregation facilitated the derivation of an empirical distribution of activations per model and data split. 
Using this empirical distribution, 
we applied various statistical techniques to quantify 
the differences in feature distributions across the different data splits for each of the models.
Furthermore, we undertook an evaluation of the normality of these distributions
using Shapiro-Wilk and Kolmogorov-Smirnov tests.
This assessment was used to determine 
whether significant deviations exist in the distribution 
of learned features across different pretraining methods.


\section{Experiments}

\paragraph{Datasets}
For pretraining, we used a publicly available dataset \cite{Guan2021} consisting of 136k organic molecules. Each molecule is represented by a single conformer generated using the Merck Molecular Force Field (MMFF94s) in the RDKit library. The initial geometry for the lowest-lying conformer was then optimized at the GFN2-xtb level of theory followed by refinement of the electronic structure with DFT  (B3LYP/def2svp). The advantage of the described dataset is several reported atomic properties: charge, electrophilicity, nucleophilicity Fukui indexes, and an NMR shielding constant. Some additional curation was carried out as described in the addendum.
Another pretraining dataset, PCQM4Mv2, consists of a single molecular property per molecule, a HOMO-LUMO gap \url{https://ogb.stanford.edu/docs/lsc/pcqm4mv2/}. It was curated under the PubChemQC project \cite{Nakata2017}. 
For the benchmarking of the obtained pretrained models, we used the absorption, distribution, metabolism, excretion, and toxicity (ADMET) group of the Therapeutics Data Commons (TDC) dataset \cite{Huang2022}, consisting of 9 regression and 13 binary classification tasks for modeling biochemical molecular properties.

\paragraph{Experiments setup}
We train our Graphormer network from scratch on the set of tasks provided by the TDC dataset.
We also pretrained the same network using as main task HOMO-LUMO gap (molecule-level)
and the atomic properties charge, electrophilicity, and nucleophilicity Fukui indexes and an NMR shielding constant (atom-level).
For atom-level pretrain, we pretrained 4 models, one per every single property.
We also pretrained a fifth model which is pretrained on 4 properties via multi-task by task-specific virtual nodes.
Main results reported in this paper for atom-level uses this last model.
Results for specific atomic property pretrained are available in the Supplementary Material.


\section{Results}
This section provides a concise summary of the key findings of this study. 
For an exhaustive elaboration of the results, please refer to the supplementary materials.

\paragraph{Pretraining on atom-level node with QM significantly improves 
performance in downstream tasks}

In Table \ref{table:Results}, we present the outcomes from 
benchmarking three distinct training approaches:
scratch, molecule-level QM pretrained, and atom-level QM pretrained
with all properties for 5 different seeds, 
as described in the guidelines provided by the TDC dataset \url{https://tdcommons.ai/benchmark/overview/}. 
We have excluded the results for atom-level pretraining on individual QM properties from this table; 
these can be found in the Supplementary Materials. 
These results show that atom-level pretraining
notably enhances model performance compared to training from scratch for 21 of the 22 datasets.

\begin{table}
    \centering
    \caption{Graphormer results}
    \label{table:Results}
    \small 
    \begin{tabular}{lllllll}
    \toprule
    & Metric & Direction & scratch & mol-level & atom-level \\
    &        &           &         &   pretrained    & pretrained \\
    &        &           &         &   HLgap    & all (4) \\
    \midrule
    caco2\_wang & MAE & $\downarrow$ & 0.48 ± 0.06 & 0.53 ± 0.02 & \textbf{0.41 ± 0.03} \\
    lipophilicity\_astrazeneca & MAE & $\downarrow$ & 0.58 ± 0.02 & 0.57 ± 0.02 & \textbf{0.42 ± 0.01} \\
    solubility\_aqsoldb & MAE & $\downarrow$ & 0.89 ± 0.04 & 0.89 ± 0.02 & \textbf{0.75 ± 0.01} \\
    ppbr\_az & MAE & $\downarrow$ & 8.38 ± 0.24 & 8.22 ± 0.23 & \textbf{7.79 ± 0.24} \\
    ld50\_zhu & MAE & $\downarrow$ & 0.61 ± 0.02 & 0.60 ± 0.03 & \textbf{0.57 ± 0.02} \\
    \midrule
    hia\_hou & ROC-AUC & $\uparrow$ & \textbf{0.96 ± 0.03} & \textbf{0.96 ± 0.02} & 0.94 ± 0.05 \\
    pgp\_broccatelli & ROC-AUC & $\uparrow$ & 0.87 ± 0.04 & 0.86 ± 0.01 & \textbf{0.89 ± 0.02} \\
    bioavailability\_ma & ROC-AUC & $\uparrow$ & 0.52 ± 0.01 & 0.55 ± 0.03 & \textbf{0.64 ± 0.05} \\
    bbb\_martins & ROC-AUC & $\uparrow$ & 0.83 ± 0.01 & 0.82 ± 0.03 & \textbf{0.88 ± 0.02} \\
    cyp3a4\_substrate\_carbonmangels & ROC-AUC & $\uparrow$ & 0.63 ± 0.07 & \textbf{0.64 ± 0.03} & \textbf{0.64 ± 0.02} \\
    ames & ROC-AUC & $\uparrow$ & 0.72 ± 0.02 & 0.73 ± 0.01 & \textbf{0.80 ± 0.01} \\
    dili & ROC-AUC & $\uparrow$ & 0.86 ± 0.02 & 0.87 ± 0.01 & \textbf{0.88 ± 0.03} \\
    herg & ROC-AUC & $\uparrow$ & \textbf{0.78 ± 0.01} & 0.76 ± 0.04 & 0.77 ± 0.06 \\
    \midrule
    vdss\_lombardo & Spearman & $\uparrow$ & 0.58 ± 0.04 & \textbf{0.59 ± 0.04} & \textbf{0.59 ± 0.03} \\
    half\_life\_obach & Spearman & $\uparrow$ & 0.39 ± 0.07 & 0.34 ± 0.07 & \textbf{0.48 ± 0.06} \\
    clearance\_microsome\_az & Spearman & $\uparrow$ & 0.49 ± 0.03 & 0.46 ± 0.03 & \textbf{0.60 ± 0.01} \\
    clearance\_hepatocyte\_az & Spearman & $\uparrow$ & 0.34 ± 0.04 & 0.31 ± 0.02 & \textbf{0.46 ± 0.03} \\
    \midrule
    cyp2d6\_veith & PR-AUC & $\uparrow$ & 0.43 ± 0.03 & 0.47 ± 0.02 & \textbf{0.61 ± 0.02} \\
    cyp3a4\_veith & PR-AUC & $\uparrow$ & 0.73 ± 0.01 & 0.74 ± 0.03 & \textbf{0.80 ± 0.03} \\
    cyp2c9\_veith & PR-AUC & $\uparrow$ & 0.63 ± 0.02 & 0.66 ± 0.03 & \textbf{0.69 ± 0.02} \\
    cyp2d6\_substrate\_carbonmangels & PR-AUC & $\uparrow$ & 0.52 ± 0.01 & 0.54 ± 0.04 & \textbf{0.58 ± 0.03} \\
    cyp2c9\_substrate\_carbonmangels & PR-AUC & $\uparrow$ & 0.35 ± 0.02 & 0.33 ± 0.03 & \textbf{0.37 ± 0.04} \\
    \bottomrule
    \end{tabular}%
\end{table}

\paragraph{Pretraining on atom-level node with QM results 
on smoother feature distribution}

\begin{figure}
    \centering
    \includegraphics[width=\textwidth]{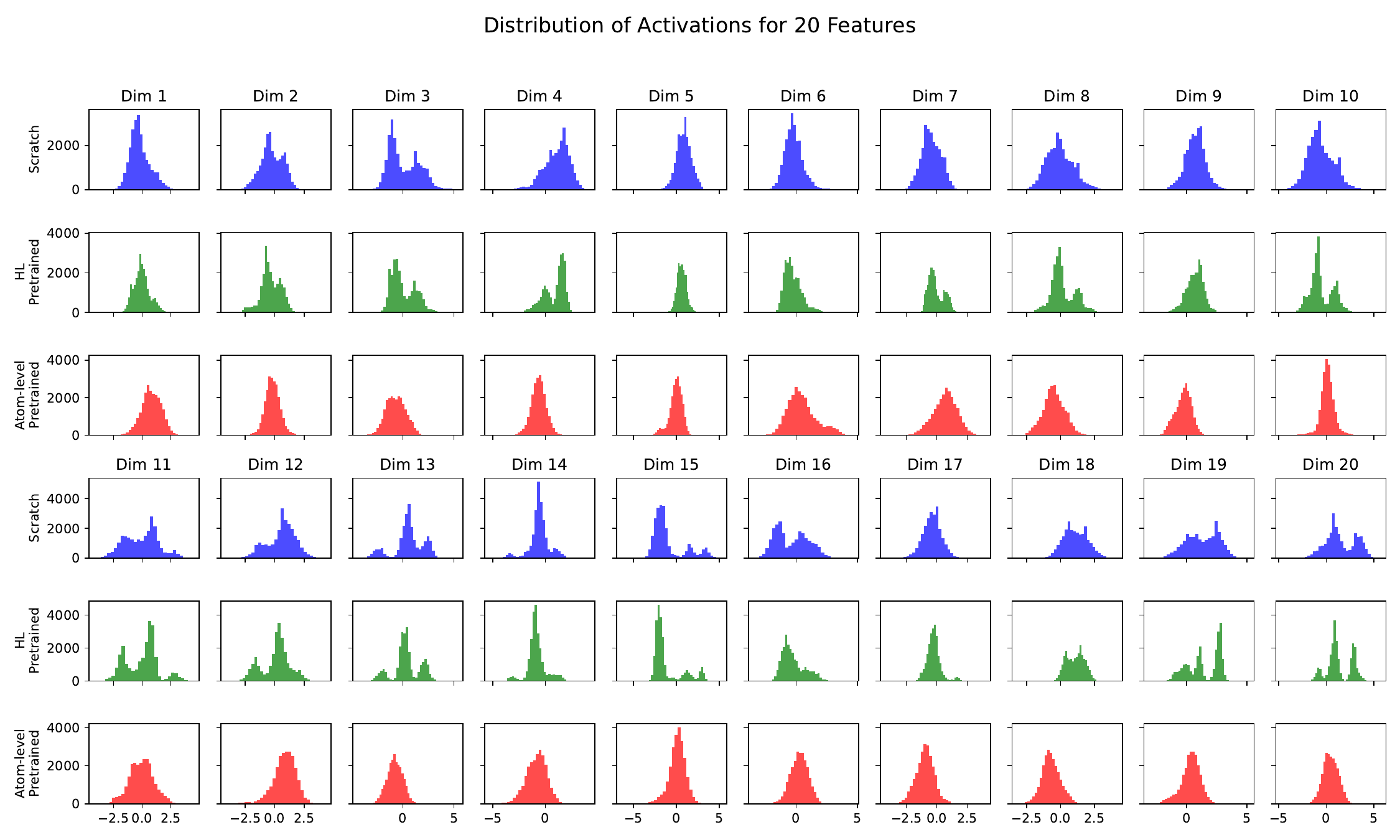}
    \caption{Distribution of first 20 features from the first layer of the Graphormer network for three different training approaches —scratch, HOMO-LUMO pretrained and atom-level pretrained— across test split of lipophilicity dataset .}
    \label{fig:distributions}
\end{figure}

We further examined the distribution of the features 
from the network's first layer for atom-level pretrained, molecular-level (HOMO-LUMO gap) and scratch models 
across different data splits and datasets. 
Figure \ref{fig:distributions} illustrates the feature distribution for the test split of one of the datasets (lipophilicity),
highlighting how scratch and molecular-level pretrain compare to the atom-level pretrained network 
(for other datasets and splits, please refer to the Supplementary Materials). 
Visually, we can observe that the atom-level pretrained method results in a more Gaussian-like distribution of the features.
We conducted a Shapiro-Wilk test for each dimension in both scratch and pretrained networks
to assess the normality of the distributions.
The average p-value for the scratch network was 3.2E-06 with a standard deviation of 5E-05, 
suggesting a significant deviation from normality. 
In contrast, the pretrained network had an average p-value of 4E-04 with a standard deviation of 0.003, 
indicating a closer approximation to a Gaussian distribution.

\paragraph{Pretraining on atom-level node with QM enhances 
robustness to input distribution shifts}

\begin{figure}
    \centering
    \includegraphics[width=\textwidth]{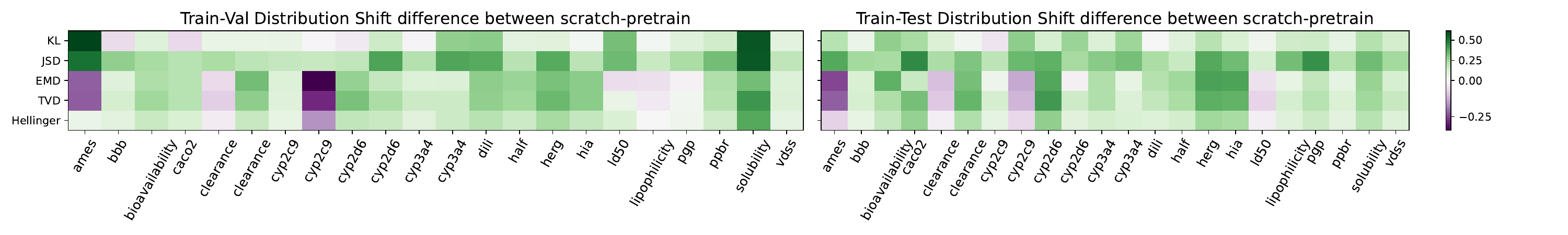}
    \caption{This heatmap illustrates the differences in distribution shifts between scratch and atom-levle pretrained networks, calculated across various feature dimensions using metrics such as Kullback-Leibler Divergence, Jensen-Shannon Divergence, Earth Mover's Distance, Total Variation Distance, and Hellinger Distance. Green hues indicate instances where the atom-levle pretrained network exhibits smaller distribution shifts compared to the scratch network, while purple hues denote the opposite. Notably, both heatmaps comparison frequently show reduced distribution shifts in atom-level pretrained networks, specially for Train-Test comparison, which likely helps to explain atom-level pretrained network's superior performance on the test data split.}
    \label{fig:distribution_shift}
\end{figure}

Pretrained networks show in general less distribution shift compared to scratch network, as we can see in Figure \ref{fig:distribution_shift}.
In this figure, we show the distribution shifts differences between scratch and pretrained networks.
As we can appreciate, in most of the cases, 
specially in the Train-Test comparison, 
atom-level pretrained networks have less distribution shift than scratch networks. 

\section{Conclusions}

In this study, we have demonstrated that 
pretraining of graph-based neural networks 
with atom-level quantum mechanics (QM) data 
significantly enhances performance on downstream tasks 
related to ADMET properties within the TDC dataset, 
as illustrated in Table \ref{table:Results}. 
Furthermore, we showed the change in the distributions of activations 
of the internal model’s features due to specific pretraining. 
After atom-level pretraining with QM data,
these distributions become more Gaussian-like, 
which is known to be conducive to better learning dynamics 
and thus improved performance (see Figure \ref{fig:distributions}). 
Moreover, our findings indicate that pretrained models 
exhibit smaller distribution shifts from training to testing datasets, 
further supporting the efficacy of QM data pretraining in enhancing model robustness (see Figure \ref{fig:distributions}). 

To our knowledge, this is the first study that elucidates 
how atom-level pretraining can optimize molecular representations
by analyzing the model's internal representation 
and robustness to distribution shifts.

\bibliographystyle{plain}
\bibliography{main}

\begin{thebibliography}{10}

\bibitem{Bajorath:09}
J.~Bajorath, L.~Peltason, M.~Wawer, R.~Guha, M.~S. Lajiness, and J.~H.~Van Drie.
\newblock Navigating structure-activity landscapes.
\newblock {\em Drug Discov Today}, 14(13-14):698--705, 2009.

\bibitem{Baptista:22}
D.~Baptista, J.~Correia, B.~Pereira, and M.~Rocha.
\newblock Evaluating molecular representations in machine learning models for drug response prediction and interpretability.
\newblock {\em J Integr Bioinform}, 19(3):20220006, 2022.

\bibitem{Bishop:06}
C.~M. Bishop.
\newblock {\em Pattern Recognition and Machine Learning}.
\newblock Springer, 2006.

\bibitem{Cai:23}
C.~Cai, T.~S. Hy, R.~Yu, and Y.~Wang.
\newblock On the connection between mpnn and graph transformer.
\newblock In {\em Proceedings of the 40th International Conference on Machine Learning}, page 138. JMLR.org, 2023.

\bibitem{Cao:23}
Z.~Cao, S.~Sciabola, and Y.~Wang.
\newblock Large-scale pretraining improves sample efficiency of active learning based molecule virtual screening.
\newblock In {\em The NeurIPS 2023 Workshop on New Frontiers of AI for Drug Discovery and Development (AI4D3 2023)}, 2023.

\bibitem{Caruana:97}
R.~Caruana.
\newblock Multitask learning.
\newblock {\em Machine Learning}, 28(1):41--75, 1997.

\bibitem{Chithrananda:20}
S.~Chithrananda, G.~Grand, and B.~Ramsundar.
\newblock Chemberta: Large-scale self-supervised pretraining for molecular property prediction, 2020.

\bibitem{Cho:23}
S.~Cho, D.-W. Jeong, S.~M. Ko, J.~Kim, S.~Han, S.~Hong, H.~Lee, and M.~Lee.
\newblock 3d denoisers are good 2d teachers: Molecular pretraining via denoising and cross-modal distillation, 2023.

\bibitem{Cruz-Monteagudo:14}
M.~Cruz-Monteagudo, J.~L. Medina-Franco, Y.~Pérez-Castillo, O.~Nicolotti, M.~N. D.~S. Cordeiro, and F.~Borges.
\newblock Activity cliffs in drug discovery: Dr jekyll or mr hyde?
\newblock {\em Drug Discovery Today}, 19(8):1069--1080, 2014.

\bibitem{Deng:23}
J.~Deng, Z.~Yang, and H.~Wang et~al.
\newblock A systematic study of key elements underlying molecular property prediction.
\newblock {\em Nat Commun}, 14:6395, 2023.

\bibitem{Devlin:19}
J.~Devlin, M.~W. Chang, K.~Lee, and K.~Toutanova.
\newblock Bert: Pre-training of deep bidirectional transformers for language understanding.
\newblock In {\em Proc. Conference of the North American Chapter of the Association for Computational Linguistics: Human Language Technologies (NAACL-HLT)}, pages 4171--4186, 2019.

\bibitem{Goodfellow-et-al-2016}
Ian Goodfellow, Yoshua Bengio, and Aaron Courville.
\newblock {\em Deep Learning}.
\newblock MIT Press, 2016.
\newblock \url{http://www.deeplearningbook.org}.

\bibitem{Guan2021}
Yanfei Guan, Connor~W. Coley, Haoyang Wu, Duminda Ranasinghe, Esther Heid, Thomas~J. Struble, Lagnajit Pattanaik, William~H. Green, and Klavs~F. Jensen.
\newblock {Regio-selectivity prediction with a machine-learned reaction representation and on-the-fly quantum mechanical descriptors}.
\newblock {\em Chemical Science}, 12(6):2198--2208, 2021.

\bibitem{Hu:20}
W.~Hu, B.~Liu, J.~Gomes, M.~Zitnik, P.~Liang, V.~Pande, and J.~Leskovec.
\newblock Strategies for pre-training graph neural networks, 2020.

\bibitem{Huang2022}
Kexin Huang, Tianfan Fu, Wenhao Gao, Yue Zhao, Yusuf Roohani, Jure Leskovec, Connor~W. Coley, Cao Xiao, Jimeng Sun, and Marinka Zitnik.
\newblock {Artificial intelligence foundation for therapeutic science}.
\newblock {\em Nature Chemical Biology}, 18(10):1033--1036, 2022.

\bibitem{Khosla:20}
Prannay Khosla, Piotr Teterwak, Chen Wang, Aaron Sarna, Yonglong Tian, Phillip Isola, Aaron Maschinot, Ce~Liu, and Dilip Krishnan.
\newblock Supervised contrastive learning.
\newblock In H.~Larochelle, M.~Ranzato, R.~Hadsell, M.F. Balcan, and H.~Lin, editors, {\em Advances in Neural Information Processing Systems}, volume~33, pages 18661--18673. Curran Associates, Inc., 2020.

\bibitem{Le:22}
T.~Le, F.~Noé, and D.~Clevert.
\newblock Equivariant graph attention networks for molecular property prediction.
\newblock In {\em Proceedings of the Learning on Graphs Conference 2022}. LoG Conference, 2022.

\bibitem{Le-Khac:20}
Phuc~H. Le-Khac, Graham Healy, and Alan~F. Smeaton.
\newblock Contrastive representation learning: A framework and review.
\newblock {\em IEEE Access}, 8:193907--193934, 2020.

\bibitem{Li:20}
X.~Li and D.~Fourches.
\newblock Inductive transfer learning for molecular activity prediction: Next-gen qsar models with molpmofit.
\newblock {\em J Cheminform}, page~27, 2020.

\bibitem{Nakata2017}
Maho Nakata and Tomomi Shimazaki.
\newblock {PubChemQC Project: A Large-Scale First-Principles Electronic Structure Database for Data-Driven Chemistry}.
\newblock {\em Journal of Chemical Information and Modeling}, 57(6):1300--1308, 2017.

\bibitem{Nugmanov:22}
R.~Nugmanov, N.~Dyubankova, A.~Gedich, and J.~K. Wegner.
\newblock Bidirectional graphormer for reactivity understanding: Neural network trained to reaction atom-to-atom mapping task.
\newblock {\em Journal of Chemical Information and Modeling}, 62(14):3307--3315, 2022.

\bibitem{Peltason:11}
L.~Peltason and J.~Bajorath.
\newblock {\em Computational Analysis of Activity and Selectivity Cliffs}, volume 672, pages 119--132.
\newblock Humana Press, 2011.

\bibitem{Press:21}
O.~Press, N.~A. Smith, and M.~Lewis.
\newblock Train short, test long: Attention with linear biases enables input length extrapolation.
\newblock In {\em Proc. Annual Meeting of the Association for Computational Linguistics}. ACL, 2021.

\bibitem{Raffel:19}
C.~Raffel, N.~Shazeer, A.~Roberts, K.~Lee, S.~Narang, M.~Matena, Y.~Zhou, W.~Li, and P.~J. Liu.
\newblock Exploring the limits of transfer learning with a unified text-to-text transformer.
\newblock In {\em Proc. Annual Conference on Neural Information Processing Systems (NeurIPS)}, 2019.

\bibitem{Ramsauer:21}
Hubert Ramsauer, Bernhard Schäfl, Johannes Lehner, Philipp Seidl, Michael Widrich, Thomas Adler, Lukas Gruber, Markus Holzleitner, Milena Pavlović, Geir~Kjetil Sandve, Victor Greiff, David Kreil, Michael Kopp, Günter Klambauer, Johannes Brandstetter, and Sepp Hochreiter.
\newblock Hopfield networks is all you need.
\newblock In {\em International Conference on Learning Representations}, 2021.

\bibitem{Robinson:20}
M.~C. Robinson, R.~C. Glen, and A.~A. Lee.
\newblock Validating the validation: reanalyzing a large-scale comparison of deep learning and machine learning models for bioactivity prediction.
\newblock {\em J Comput Aided Mol Des}, 34:717--730, 2020.

\bibitem{Rosenstein:05}
M.~T. Rosenstein, Z.~Marx, L.~P. Kaelbling, and T.~G. Dietterich.
\newblock To transfer or not to transfer.
\newblock In {\em Advances in Neural Information Processing Systems (NeurIPS), Workshop on transfer learning}, volume 898, pages 1--4, 2005.

\bibitem{Sanchez-Fernandez:23}
A.~Sanchez-Fernandez, E.~Rumetshofer, and S.~Hochreiter et~al.
\newblock Cloome: contrastive learning unlocks bioimaging databases for queries with chemical structures.
\newblock {\em Nat Commun}, 14:7339, 2023.

\bibitem{Satorras:21}
V.~Garcia Satorras, E.~Hoogeboom, and M.~Welling.
\newblock E(n) equivariant graph neural networks.
\newblock In {\em Proceedings of the International Conference on Learning Representations (ICLR)}, 2021.

\bibitem{Schimunek:23}
Johannes Schimunek, Philipp Seidl, Lukas Friedrich, Daniel Kuhn, Friedrich Rippmann, Sepp Hochreiter, and Günter Klambauer.
\newblock Context-enriched molecule representations improve few-shot drug discovery.
\newblock In {\em The Eleventh International Conference on Learning Representations}, 2023.

\bibitem{Seidl:24}
Philipp Seidl.
\newblock {\em Multimodal Contrastive Learning for Drug Discovery}.
\newblock Dissertation (phd), Johannes Kepler University Linz, Linz, 2024.
\newblock Includes illustrations.

\bibitem{Stumpfe:12}
D.~Stumpfe and J.~Bajorath.
\newblock Exploring activity cliffs in medicinal chemistry miniperspective.
\newblock {\em Journal of Medicinal Chemistry}, 55(7):2932--2942, 2012.

\bibitem{Stark:23}
H.~Stärk, D.~Beaini, G.~Corso, P.~Tossou, C.~Dallago, S.~Günnemann, and P.~Lió.
\newblock 3d infomax improves gnns for molecular property prediction.
\newblock In {\em Proceedings of the 39th International Conference on Machine Learning (ICML)}. PMLR, 2023.

\bibitem{Tossou:23}
P.~Tossou, C.~Wognum, M.~Craig, M.~H, and E.~Noutahi.
\newblock Real-world molecular out-of-distribution: Specification and investigation.
\newblock {\em ChemRxiv}, 2023.
\newblock This content is a preprint and has not been peer-reviewed.

\bibitem{vanTilborg:22}
D.~van Tilborg, A.~Alenicheva, and F.~Grisoni.
\newblock Exposing the limitations of molecular machine learning with activity cliffs.
\newblock {\em Journal of Chemical Information and Modeling}, 62(23):5938--5951, 2022.

\bibitem{Wang:19}
S.~Wang, Y.~Guo, Y.~Wang, H.~Sun, and J.~Huang.
\newblock Smiles-bert: Large scale unsupervised pre-training for molecular property prediction.
\newblock In {\em Proceedings of the 10th ACM International Conference on Bioinformatics, Computational Biology and Health Informatics (ACM-BCB)}, pages 429--436, Niagara Falls, NY, USA, 2019. ACM.

\bibitem{Wang:22}
Yuyang Wang, Jianren Wang, Zhonglin Cao, and Amir Barati~Farimani et~al.
\newblock Molecular contrastive learning of representations via graph neural networks.
\newblock {\em Nat Mach Intell}, 4:279--287, 2022.

\bibitem{Xia:23}
J.~Xia, L.~Zhang, X.~Zhu, Y.~Liu, Z.~Gao, B.~Hu, C.~Tan, J.~Zheng, S.~Li, and S.~Z. Li.
\newblock Understanding the limitations of deep models for molecular property prediction: Insights and solutions.
\newblock In {\em Advances in Neural Information Processing Systems}, volume~36, pages 64774--64792. Curran Associates, Inc., 2023.

\bibitem{Xia:23b}
J.~Xia, Y.~Zhu, Y.~Du, and S.~Z. Li.
\newblock A systematic survey of chemical pre-trained models.
\newblock In {\em Proceedings of the Thirty-Second International Joint Conference on Artificial Intelligence (IJCAI-23)}. IJCAI, 2023.

\bibitem{Xu:24}
L.~Xu, L.~Xia, S.~Pan, and Z.~Li.
\newblock Triple generative self-supervised learning method for molecular property prediction.
\newblock {\em Int. J. Mol. Sci.}, 25(7):3794, 2024.

\bibitem{Ying:21}
C.~Ying, T.~Cai, S.~Luo, S.~Zheng, G.~Ke, D.~He, Y.~Shen, and T.-Y. Liu.
\newblock Do transformers really perform badly for graph representation?
\newblock In M.~Ranzato, A.~Beygelzimer, Y.~Dauphin, P.~S. Liang, and J.~Wortman Vaughan, editors, {\em Advances in Neural Information Processing Systems}, volume~34, pages 28877--28888. Curran Associates, Inc., 2021.

\bibitem{Zaidi:22}
S.~Zaidi, M.~Schaarschmidt, J.~Martens, H.~Kim, Y.~W. Teh, A.~Sanchez-Gonzalez, P.~Battaglia, R.~Pascanu, and J.~Godwin.
\newblock Pre-training via denoising for molecular property prediction, 2022.

\bibitem{Zhang:24}
Ruochi Zhang, Chao Wu, Qian Yang, Chang Liu, Yan Wang, Kewei Li, Lan Huang, and Fengfeng Zhou.
\newblock Molfescue: enhancing molecular property prediction in data-limited and imbalanced contexts using few-shot and contrastive learning.
\newblock {\em Bioinformatics}, 40:btae118, 2024.

\bibitem{Zhao:23}
H.~Zhao, S.~Liu, C.~Ma, H.~Xu, J.~Fu, Z.-H. Deng, L.~Kong, and Q.~Liu.
\newblock Gimlet: A unified graph-text model for instruction-based molecule zero-shot learning, 2023.

\end{thebibliography}

\clearpage
\appendix
\section{Supplementary Materials}

\subsection{Full table results}

\subsection{Distribution of activations of features for scratch, HOMO-LUMO pretrained and Atom-level pretrained networks in TDC Lipophilicity Dataset}

\subsection{Distribution of first 20 features from the first layer of the Graphormer network for three different training approaches —scratch, HOMO-LUMO pretrained and atom-level pretrained— across test split of TDC datasets}

\clearpage

\begin{landscape}
\begin{table}
    \centering
    \caption{Graphormer results}
    \label{table:Results_extended}
    \small 
    \begin{tabular}{llllllllllll}
    \toprule
    & Metric & Direction & scratch & HLgap & atom-level &  &  & &  & mol-level \\
    &        &           &         &       & pretrained &         &     &         &        &  pretrained     \\
        &        &           &         &       & all (4) &    fukui\_e     & nmr    &  fukui\_n        & charge       &  HLgap     \\
    \midrule
    caco2\_wang & MAE & $\downarrow$ & 0.48 ± 0.06 & 0.53 ± 0.02 & \textbf{0.41 ± 0.03} & 0.45 ± 0.07 & 0.48 ± 0.06 & \textbf{0.39 ± 0.02} & 0.40 ± 0.08 & 0.53 ± 0.02 \\
    lipophilicity\_astrazeneca & MAE & $\downarrow$ & 0.58 ± 0.02 & 0.57 ± 0.02 & \textbf{0.42 ± 0.01} & 0.49 ± 0.02 & 0.46 ± 0.01 & 0.48 ± 0.01 & \textbf{0.43 ± 0.01} & 0.57 ± 0.02 \\
    solubility\_aqsoldb & MAE & $\downarrow$ & 0.89 ± 0.04 & 0.89 ± 0.02 & \textbf{0.75 ± 0.01} & 0.80 ± 0.02 & 0.78 ± 0.02 & 0.78 ± 0.02 & \textbf{0.75 ± 0.01} & 0.89 ± 0.02 \\
    ppbr\_az & MAE & $\downarrow$ & 8.38 ± 0.24 & 8.22 ± 0.23 & \textbf{7.79 ± 0.24} & 7.92 ± 0.12 & 8.02 ± 0.40 & 7.79 ± 0.28 & \textbf{7.57 ± 0.32} & 8.22 ± 0.23 \\
    ld50\_zhu & MAE & $\downarrow$ & 0.61 ± 0.02 & 0.60 ± 0.03 & \textbf{0.57 ± 0.02} & 0.60 ± 0.01 & \textbf{0.56 ± 0.02} & 0.60 ± 0.02 & 0.57 ± 0.01 & 0.60 ± 0.03 \\
    \midrule
    hia\_hou & ROC-AUC & $\uparrow$ & \textbf{0.96 ± 0.03} & \textbf{0.96 ± 0.02} & 0.94 ± 0.05 & 0.93 ± 0.03 & \textbf{0.97 ± 0.02} & 0.94 ± 0.02 & 0.95 ± 0.02 & 0.96 ± 0.02 \\
    pgp\_broccatelli & ROC-AUC & $\uparrow$ & 0.87 ± 0.04 & 0.86 ± 0.01 & \textbf{0.89 ± 0.02} & 0.89 ± 0.03 & 0.86 ± 0.03 & \textbf{0.90 ± 0.01} & 0.88 ± 0.02 & 0.86 ± 0.01 \\
    bioavailability\_ma & ROC-AUC & $\uparrow$ & 0.52 ± 0.01 & 0.55 ± 0.03 & \textbf{0.64 ± 0.05} & 0.64 ± 0.02 & 0.66 ± 0.01 & \textbf{0.69 ± 0.05} & 0.62 ± 0.07 & 0.55 ± 0.03 \\
    bbb\_martins & ROC-AUC & $\uparrow$ & 0.83 ± 0.01 & 0.82 ± 0.03 & \textbf{0.88 ± 0.02} & 0.86 ± 0.03 & 0.86 ± 0.02 & 0.85 ± 0.02 & \textbf{0.87 ± 0.01} & 0.82 ± 0.03 \\
    cyp3a4\_substrate\_carbonmangels & ROC-AUC & $\uparrow$ & 0.63 ± 0.07 & 0.64 ± 0.03 & 0.64 ± 0.02 & \textbf{0.66 ± 0.03} & 0.62 ± 0.02 & 0.61 ± 0.02 & 0.63 ± 0.02 & 0.64 ± 0.03 \\
    ames & ROC-AUC & $\uparrow$ & 0.72 ± 0.02 & 0.73 ± 0.01 & \textbf{0.80 ± 0.01} & 0.78 ± 0.02 & \textbf{0.80 ± 0.02} & 0.76 ± 0.01 & \textbf{0.80 ± 0.01} & 0.73 ± 0.01 \\
    dili & ROC-AUC & $\uparrow$ & 0.86 ± 0.02 & 0.87 ± 0.01 & \textbf{0.88 ± 0.03} & 0.86 ± 0.04 & \textbf{0.89 ± 0.03} & 0.82 ± 0.04 & 0.85 ± 0.04 & 0.87 ± 0.01 \\
    herg & ROC-AUC & $\uparrow$ & \textbf{0.78 ± 0.01} & 0.76 ± 0.04 & 0.77 ± 0.06 & 0.73 ± 0.06 & 0.77 ± 0.05 & \textbf{0.77 ± 0.02} & \textbf{0.79 ± 0.03} & 0.76 ± 0.04 \\
    \midrule
    vdss\_lombardo & Spearman & $\uparrow$ & 0.58 ± 0.04 & 0.59 ± 0.04 & 0.59 ± 0.03 & \textbf{0.64 ± 0.02} & 0.61 ± 0.04 & 0.61 ± 0.01 & \textbf{0.63 ± 0.03} & 0.59 ± 0.04 \\
    half\_life\_obach & Spearman & $\uparrow$ & 0.39 ± 0.07 & 0.34 ± 0.07 & \textbf{0.48 ± 0.06} & \textbf{0.48 ± 0.04} & 0.42 ± 0.10 & 0.48 ± 0.03 & 0.47 ± 0.04 & 0.34 ± 0.07 \\
    clearance\_microsome\_az & Spearman & $\uparrow$ & 0.49 ± 0.03 & 0.46 ± 0.03 & \textbf{0.60 ± 0.01} & 0.47 ± 0.06 & 0.57 ± 0.01 & \textbf{0.58 ± 0.02} & \textbf{0.58 ± 0.01} & 0.46 ± 0.03 \\
    clearance\_hepatocyte\_az & Spearman & $\uparrow$ & 0.34 ± 0.04 & 0.31 ± 0.02 & \textbf{0.46 ± 0.03} & 0.42 ± 0.02 & 0.44 ± 0.04 & 0.41 ± 0.02 & 0.46 ± 0.04 & 0.31 ± 0.02 \\
    \midrule
    cyp2d6\_veith & PR-AUC & $\uparrow$ & 0.43 ± 0.03 & 0.47 ± 0.02 & \textbf{0.61 ± 0.02} & 0.55 ± 0.03 & 0.56 ± 0.04 & 0.56 ± 0.02 & 0.58 ± 0.04 & 0.47 ± 0.02 \\
    cyp3a4\_veith & PR-AUC & $\uparrow$ & 0.73 ± 0.01 & 0.74 ± 0.03 & \textbf{0.80 ± 0.03} & 0.77 ± 0.03 & 0.78 ± 0.01 & \textbf{0.79 ± 0.03} & 0.76 ± 0.04 & 0.74 ± 0.03 \\
    cyp2c9\_veith & PR-AUC & $\uparrow$ & 0.63 ± 0.02 & 0.66 ± 0.03 & \textbf{0.69 ± 0.02} & 0.67 ± 0.02 & 0.69 ± 0.04 & \textbf{0.69 ± 0.01} & 0.69 ± 0.02 & 0.66 ± 0.03 \\
    cyp2d6\_substrate\_carbonmangels & PR-AUC & $\uparrow$ & 0.52 ± 0.01 & 0.54 ± 0.04 & \textbf{0.58 ± 0.03} & 0.53 ± 0.06 & \textbf{0.64 ± 0.06} & 0.57 ± 0.04 & 0.63 ± 0.03 & 0.54 ± 0.04 \\
    cyp2c9\_substrate\_carbonmangels & PR-AUC & $\uparrow$ & 0.35 ± 0.02 & 0.33 ± 0.03 & \textbf{0.37 ± 0.04} & 0.32 ± 0.04 & 0.34 ± 0.03 & \textbf{0.37 ± 0.04} & 0.36 ± 0.04 & 0.33 ± 0.03 \\
    \bottomrule
    \end{tabular}%
\end{table}
\end{landscape}

\begin{landscape}
\begin{figure}
    \centering
    \includegraphics[width=\paperwidth, height=\paperheight, keepaspectratio]{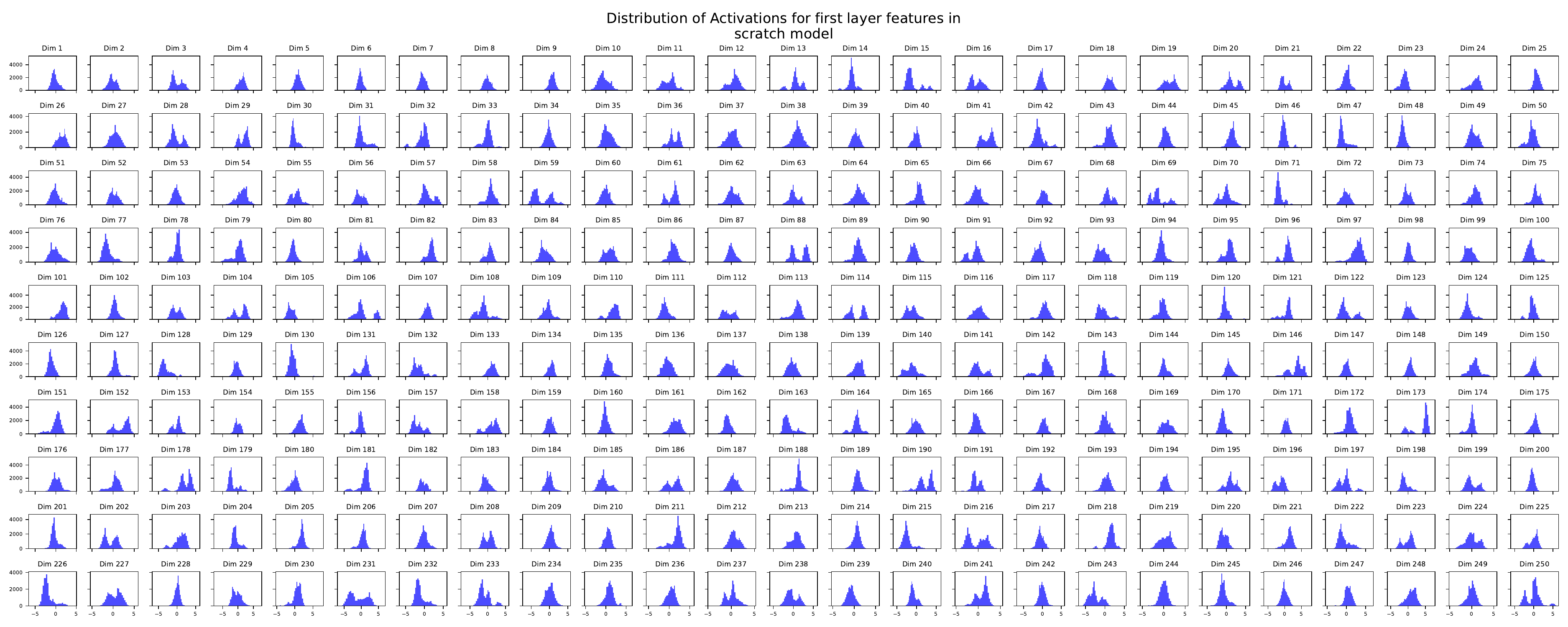}
    \caption{Distribution plots Scratch for Lipophilicity test set}
    \label{fig:dit_scratch_1}
\end{figure}
\end{landscape}


\begin{landscape}
\begin{figure}
    \centering
    \includegraphics[width=\paperwidth, height=\paperheight, keepaspectratio]{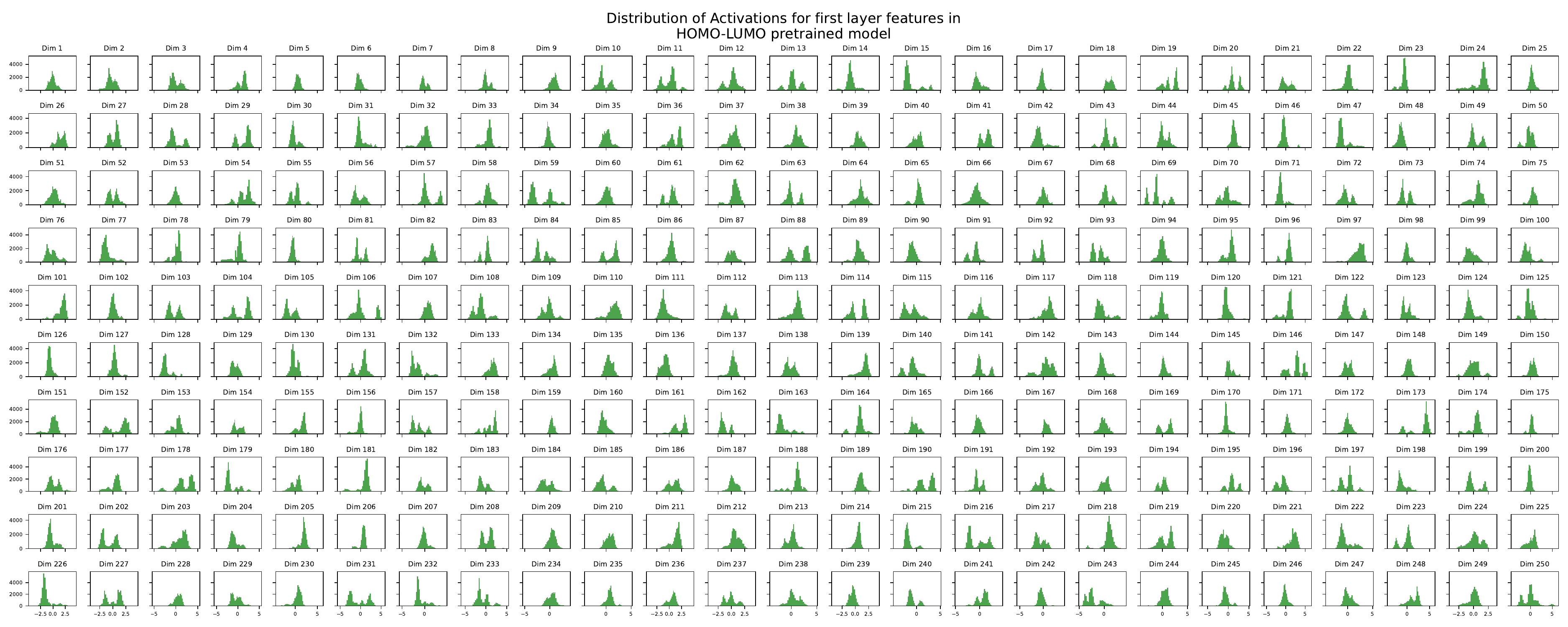}
    \caption{Distribution plots HOMO-LUMO pretrained for Lipophilicity test set}
    \label{fig:dit_HL}
\end{figure}
\end{landscape}


\begin{landscape}
\begin{figure}
    \centering
    \includegraphics[width=\paperwidth, height=\paperheight, keepaspectratio]{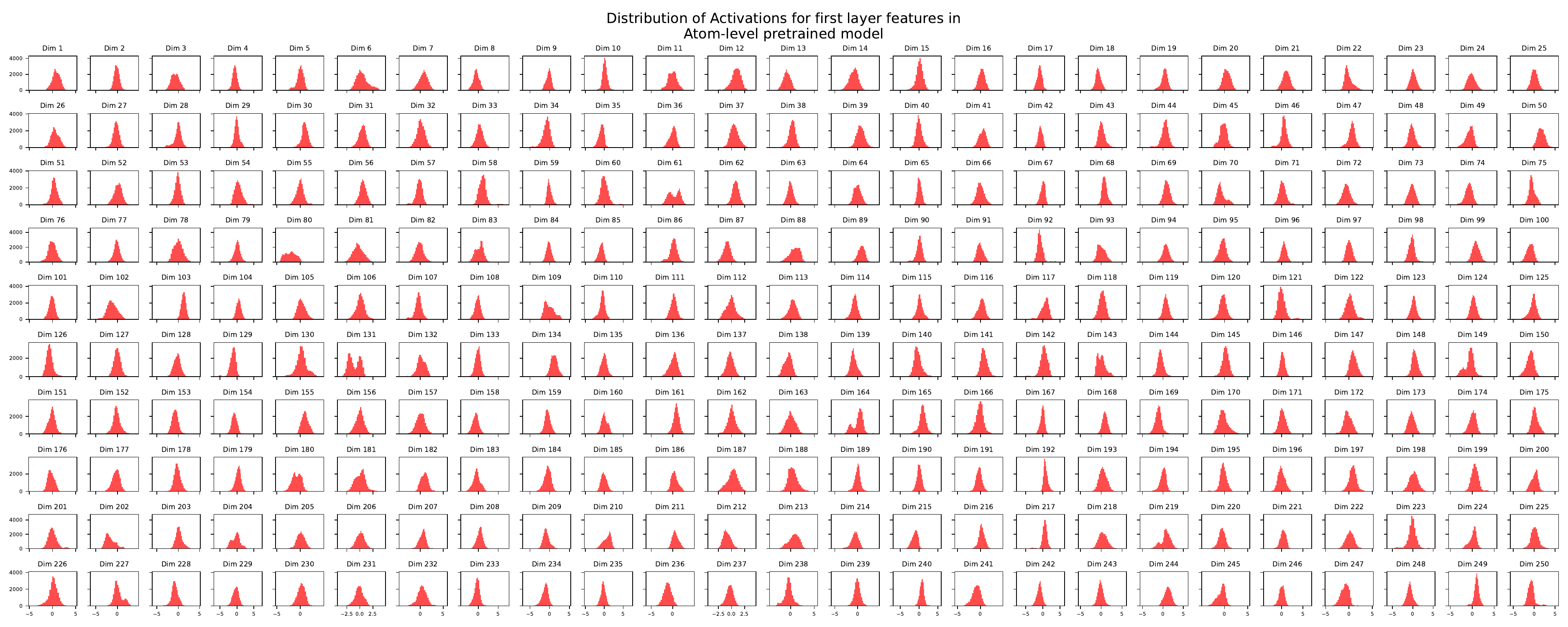}
    \caption{Distribution plots Atom-level pretrained for Lipophilicity test set}
    \label{fig:dit_atom}
\end{figure}
\end{landscape}

\begin{figure}
    \centering
    \includegraphics[width=\textwidth]{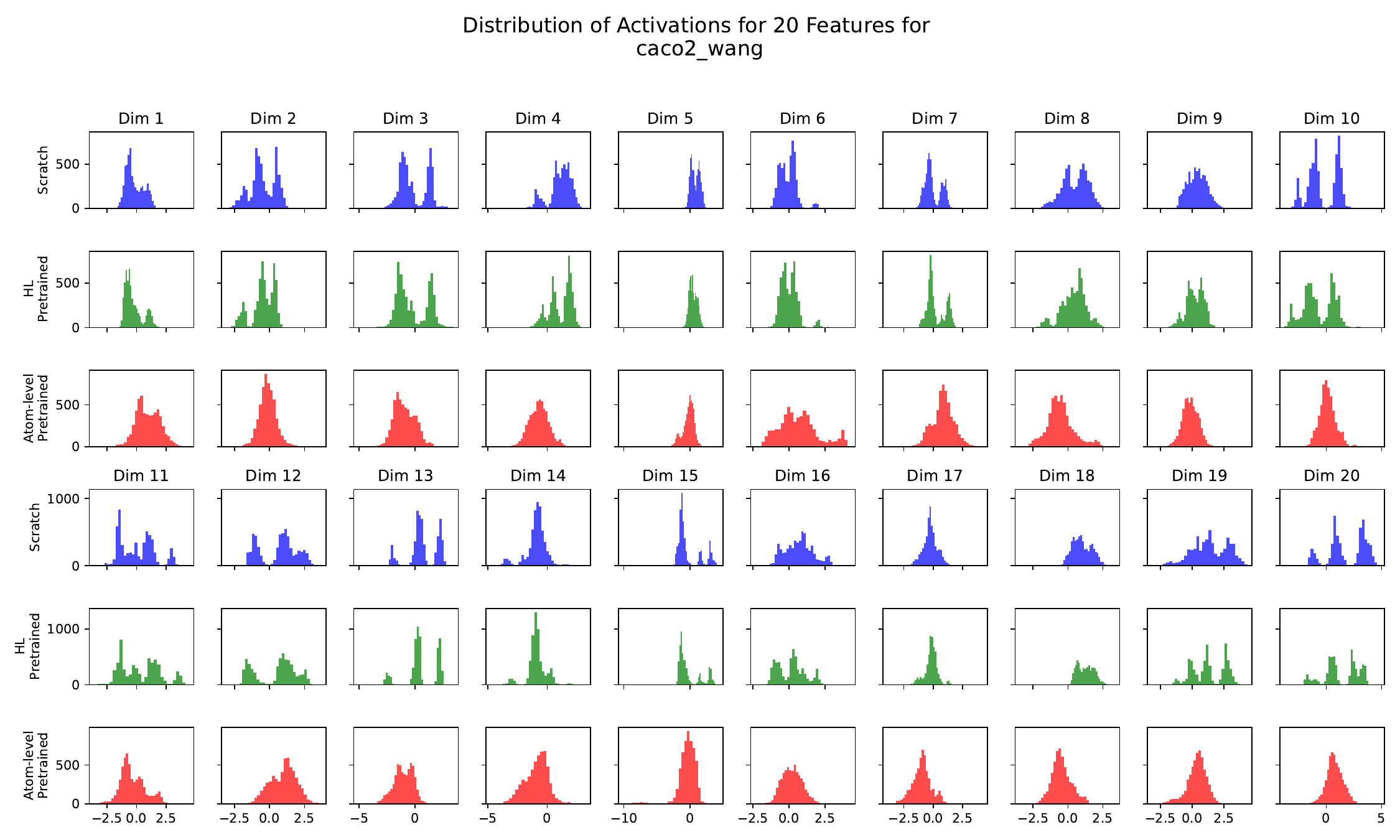}
    \caption{Distribution of first 20 features from the first layer of the Graphormer network for three different training approaches —scratch, HOMO-LUMO pretrained and atom-level pretrained— across test split of caco2 wang dataset.}
    \label{fig:distributions_caco2_wang}
\end{figure}

\begin{figure}
    \centering
    \includegraphics[width=\textwidth]{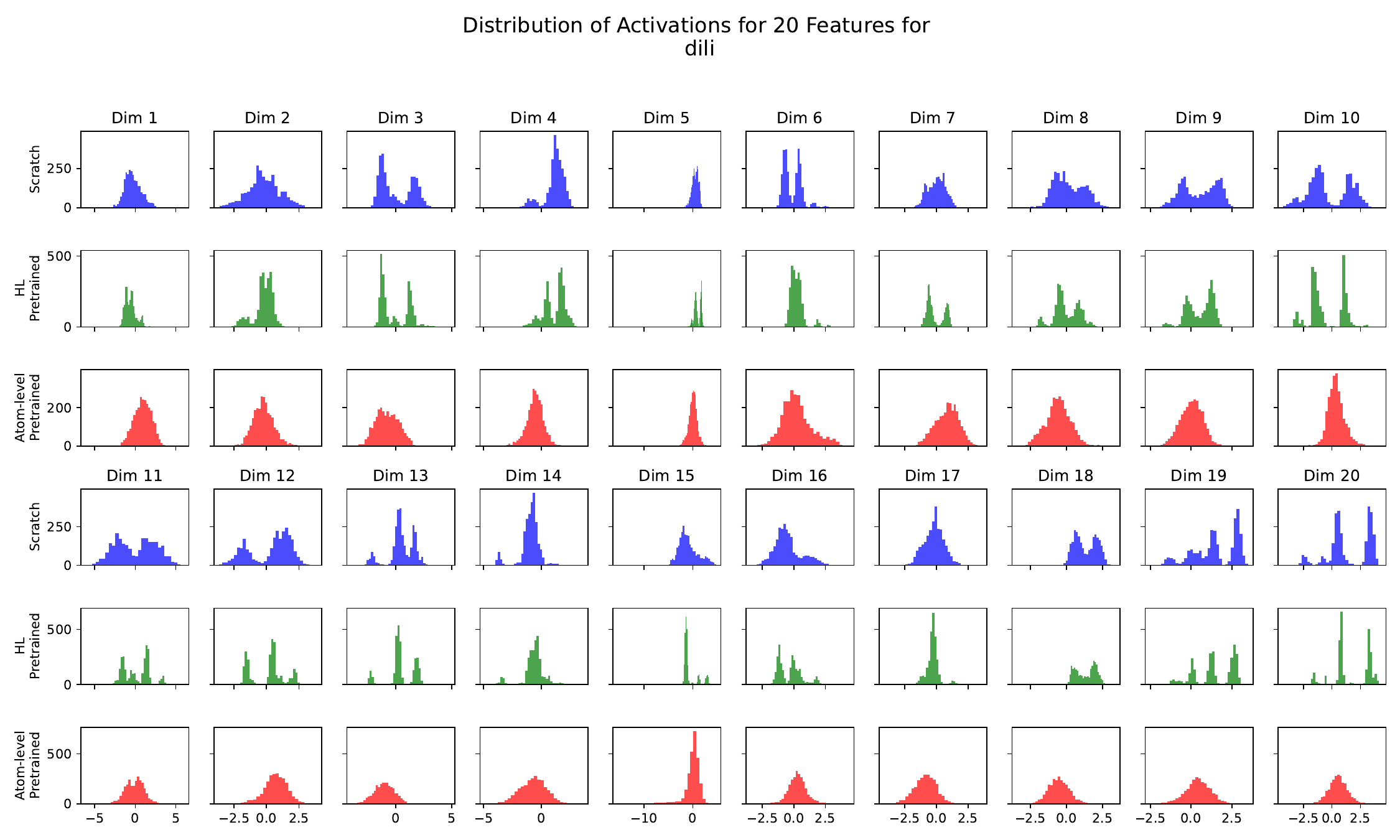}
    \caption{Distribution of first 20 features from the first layer of the Graphormer network for three different training approaches —scratch, HOMO-LUMO pretrained and atom-level pretrained— across test split of dili dataset.}
    \label{fig:distributions_dili}
\end{figure}

\begin{figure}
    \centering
    \includegraphics[width=\textwidth]{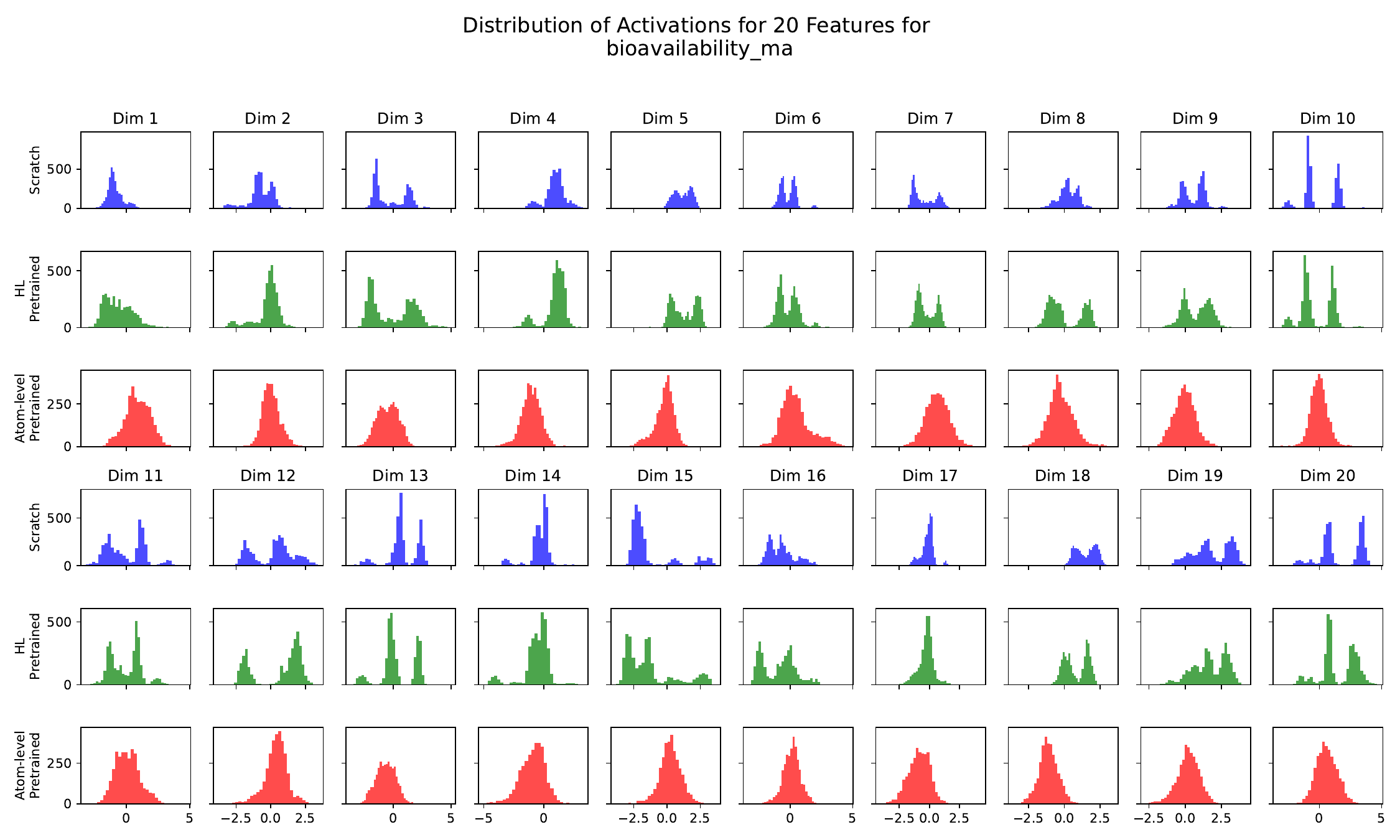}
    \caption{Distribution of first 20 features from the first layer of the Graphormer network for three different training approaches —scratch, HOMO-LUMO pretrained and atom-level pretrained— across test split of bioavailability ma dataset.}
    \label{fig:distributions_bioavailability_ma}
\end{figure}

\begin{figure}
    \centering
    \includegraphics[width=\textwidth]{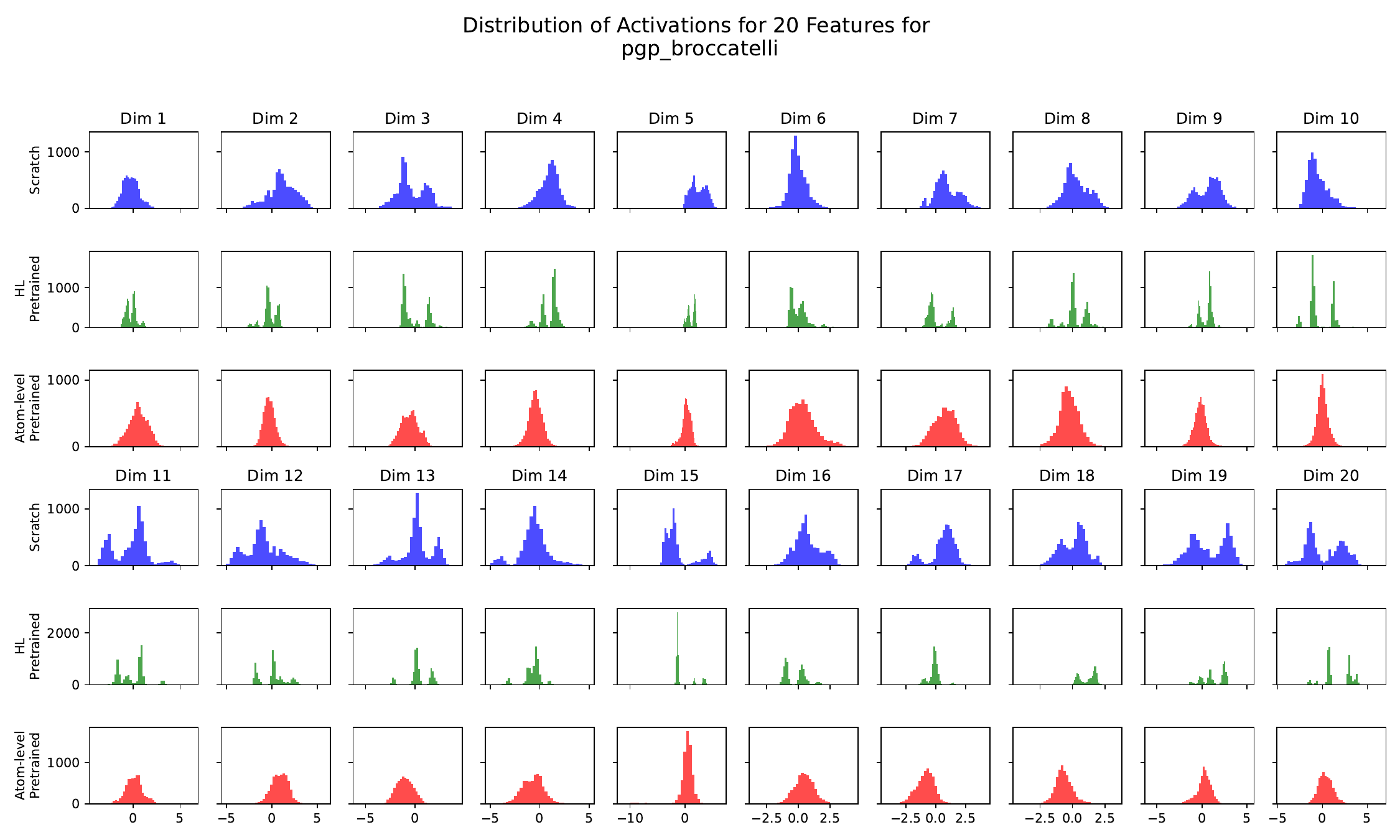}
    \caption{Distribution of first 20 features from the first layer of the Graphormer network for three different training approaches —scratch, HOMO-LUMO pretrained and atom-level pretrained— across test split of pgp broccatelli dataset.}
    \label{fig:distributions_pgp_broccatelli}
\end{figure}

\begin{figure}
    \centering
    \includegraphics[width=\textwidth]{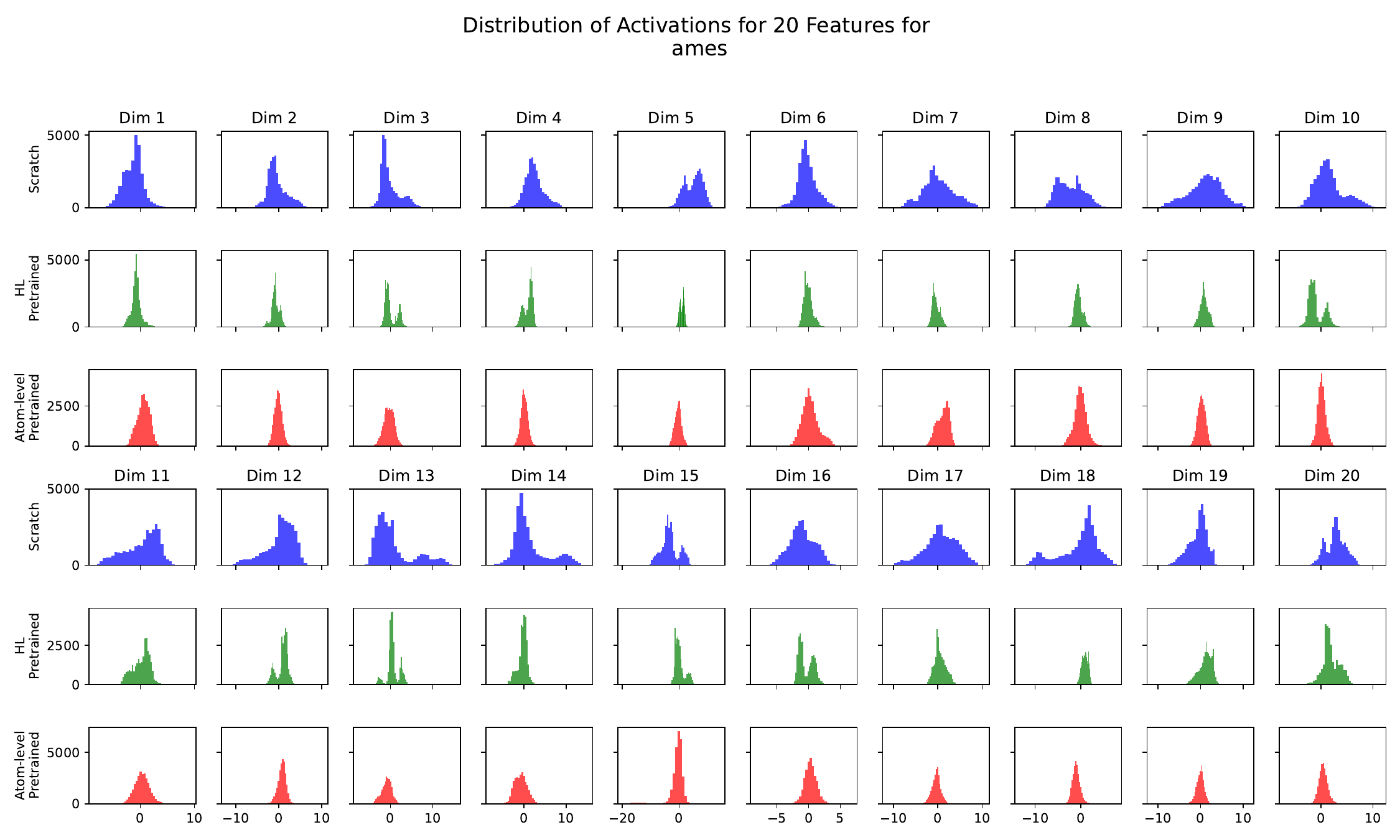}
    \caption{Distribution of first 20 features from the first layer of the Graphormer network for three different training approaches —scratch, HOMO-LUMO pretrained and atom-level pretrained— across test split of ames dataset.}
    \label{fig:distributions_ames}
\end{figure}

\begin{figure}
    \centering
    \includegraphics[width=\textwidth]{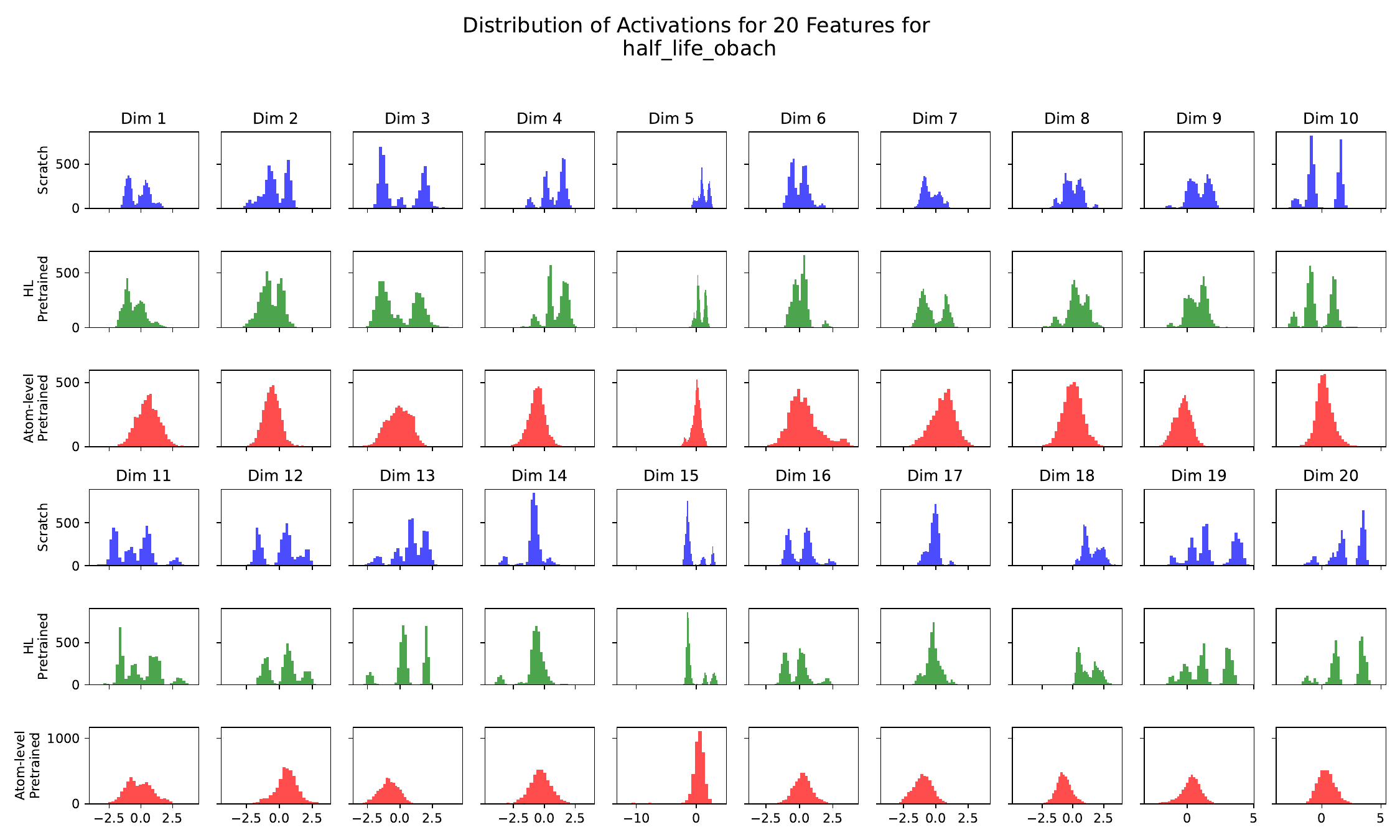}
    \caption{Distribution of first 20 features from the first layer of the Graphormer network for three different training approaches —scratch, HOMO-LUMO pretrained and atom-level pretrained— across test split of half life obach dataset.}
    \label{fig:distributions_half_life_obach}
\end{figure}

\begin{figure}
    \centering
    \includegraphics[width=\textwidth]{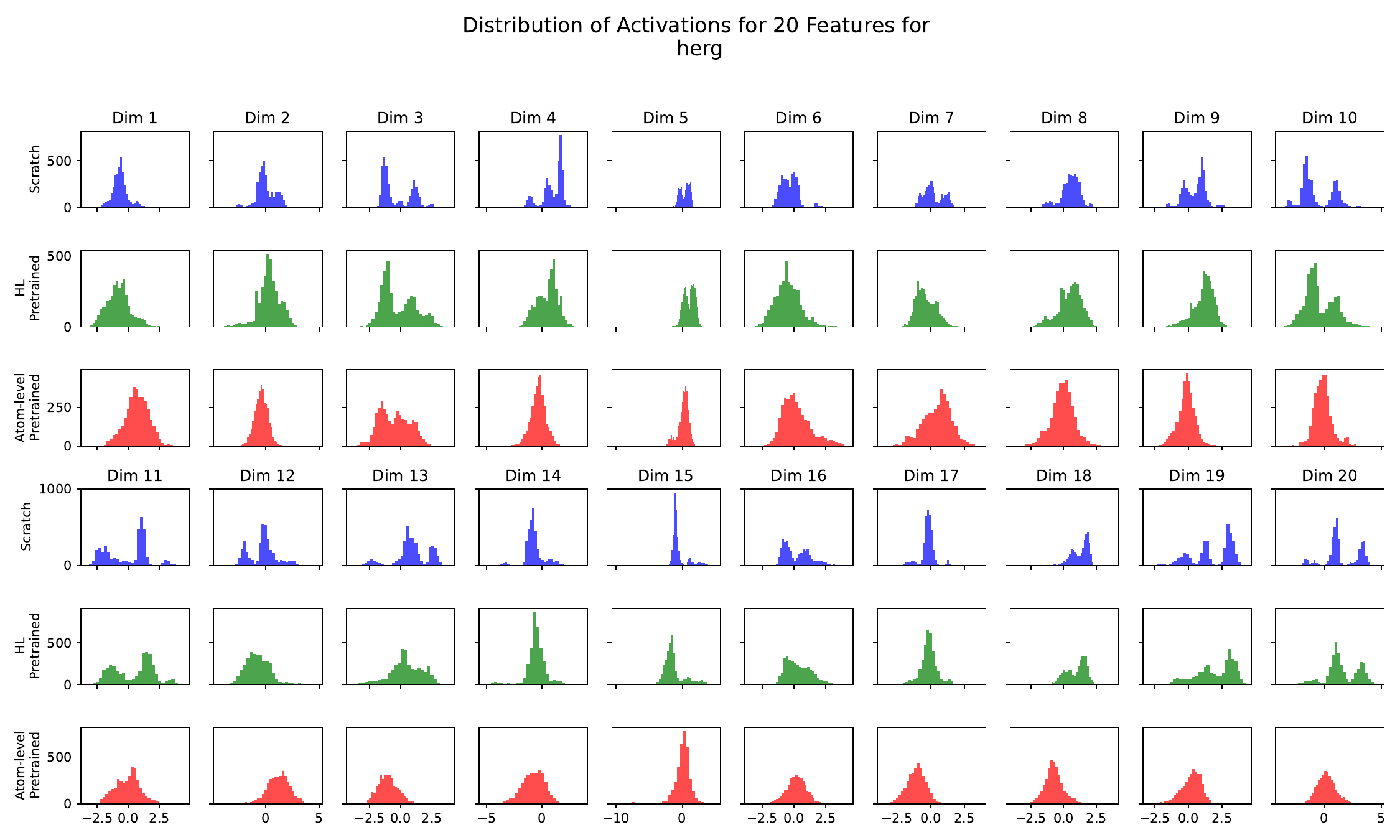}
    \caption{Distribution of first 20 features from the first layer of the Graphormer network for three different training approaches —scratch, HOMO-LUMO pretrained and atom-level pretrained— across test split of herg dataset.}
    \label{fig:distributions_herg}
\end{figure}

\begin{figure}
    \centering
    \includegraphics[width=\textwidth]{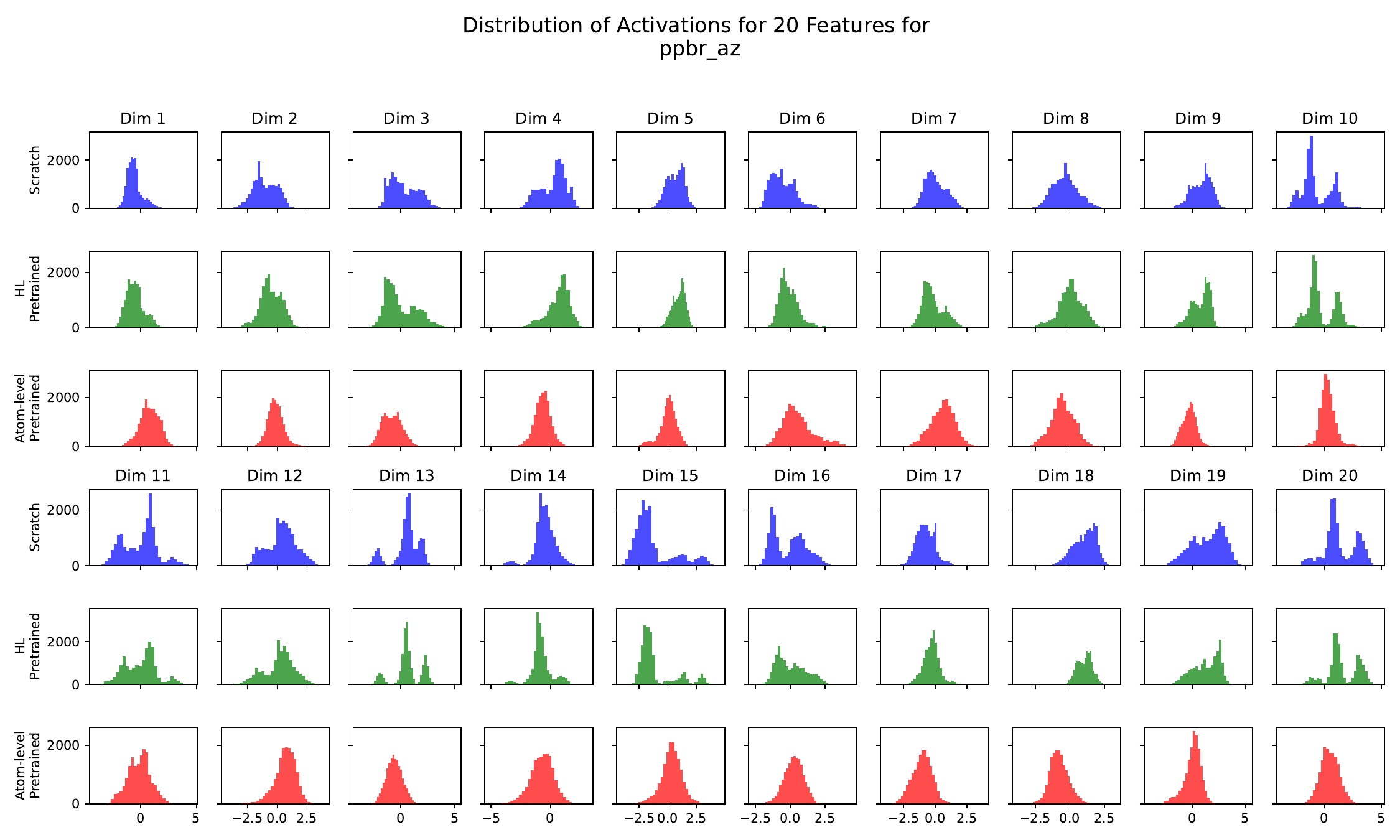}
    \caption{Distribution of first 20 features from the first layer of the Graphormer network for three different training approaches —scratch, HOMO-LUMO pretrained and atom-level pretrained— across test split of ppbr az dataset.}
    \label{fig:distributions_ppbr_az}
\end{figure}

\begin{figure}
    \centering
    \includegraphics[width=\textwidth]{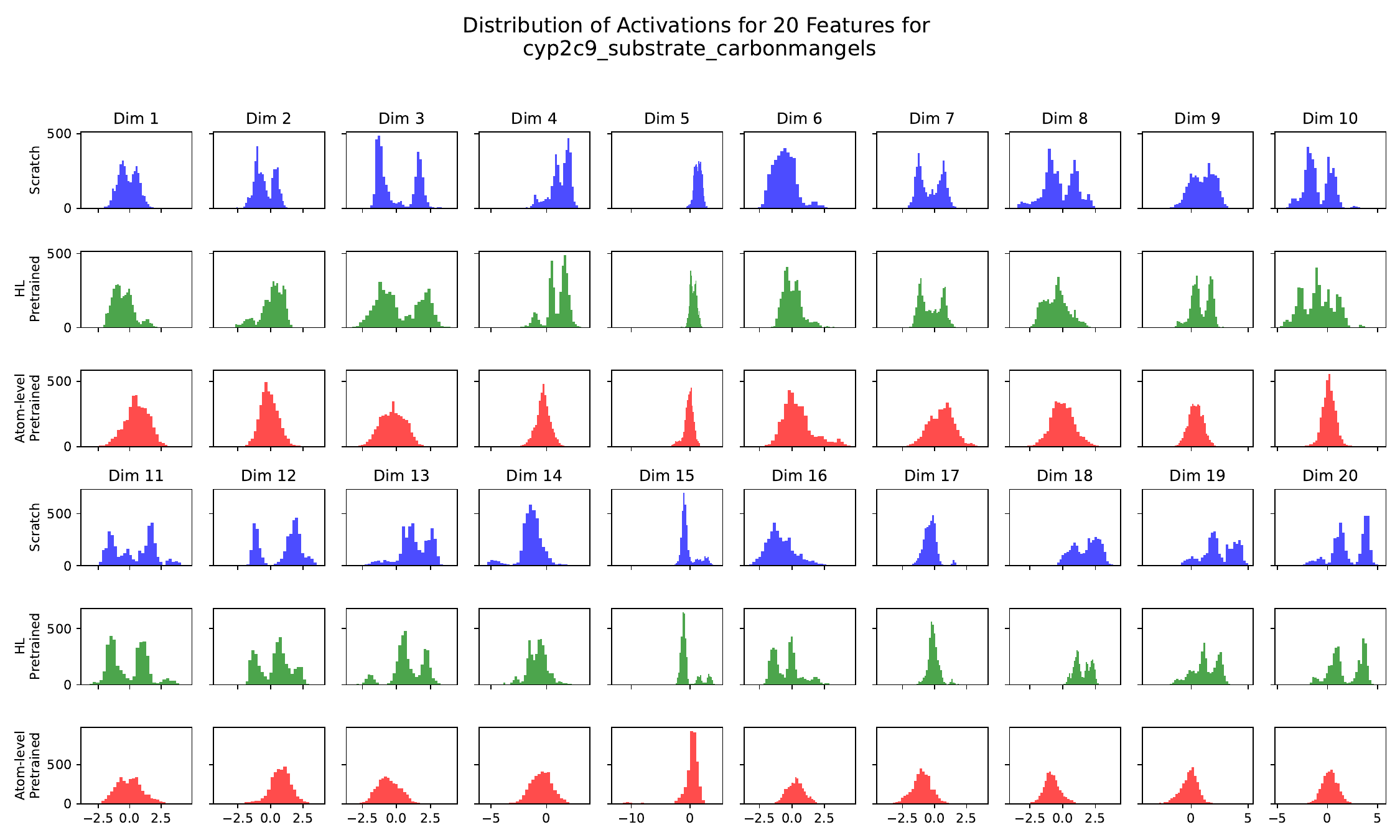}
    \caption{Distribution of first 20 features from the first layer of the Graphormer network for three different training approaches —scratch, HOMO-LUMO pretrained and atom-level pretrained— across test split of cyp2c9 substrate carbonmangels dataset.}
    \label{fig:distributions_cyp2c9_substrate_carbonmangels}
\end{figure}

\begin{figure}
    \centering
    \includegraphics[width=\textwidth]{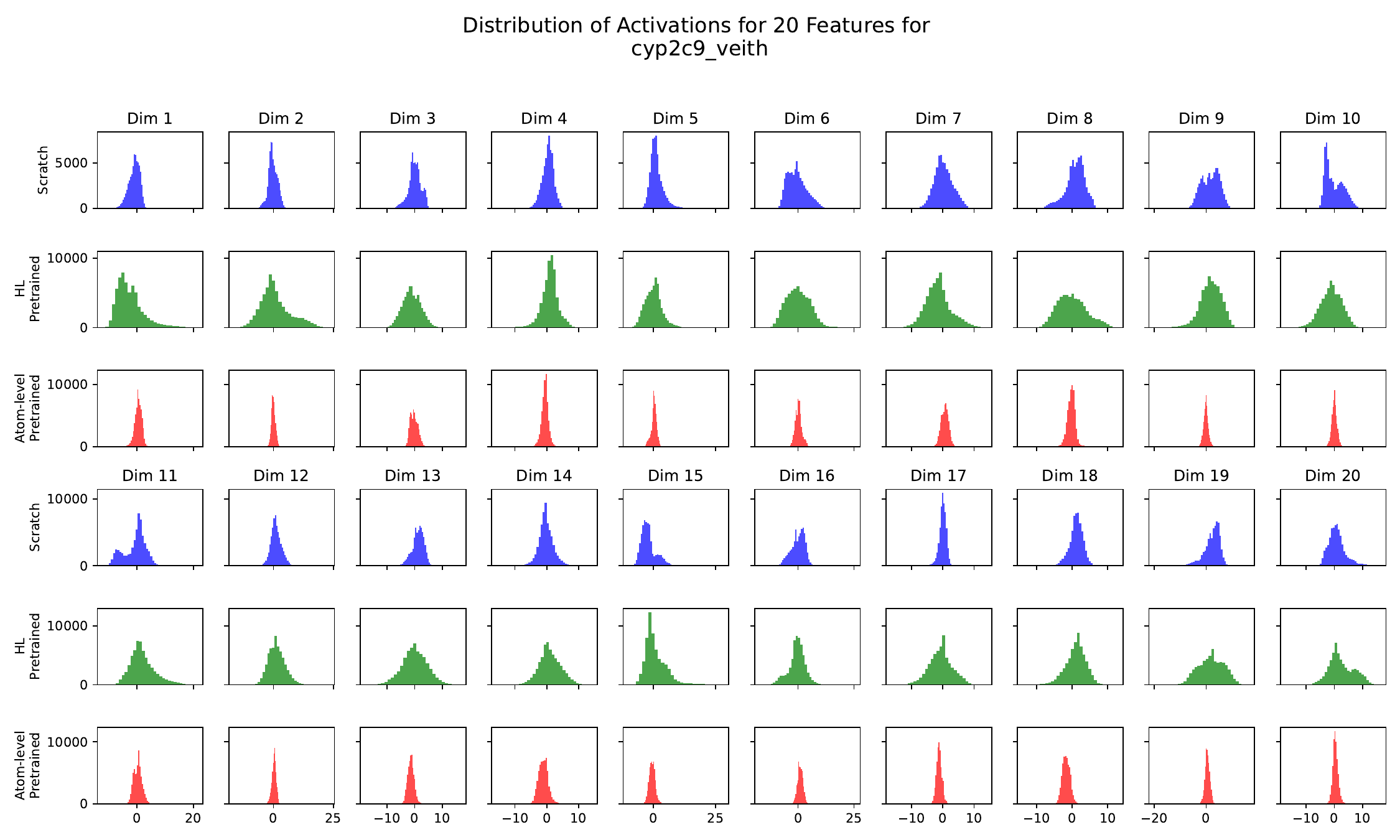}
    \caption{Distribution of first 20 features from the first layer of the Graphormer network for three different training approaches —scratch, HOMO-LUMO pretrained and atom-level pretrained— across test split of cyp2c9 veith dataset.}
    \label{fig:distributions_cyp2c9_veith}
\end{figure}

\begin{figure}
    \centering
    \includegraphics[width=\textwidth]{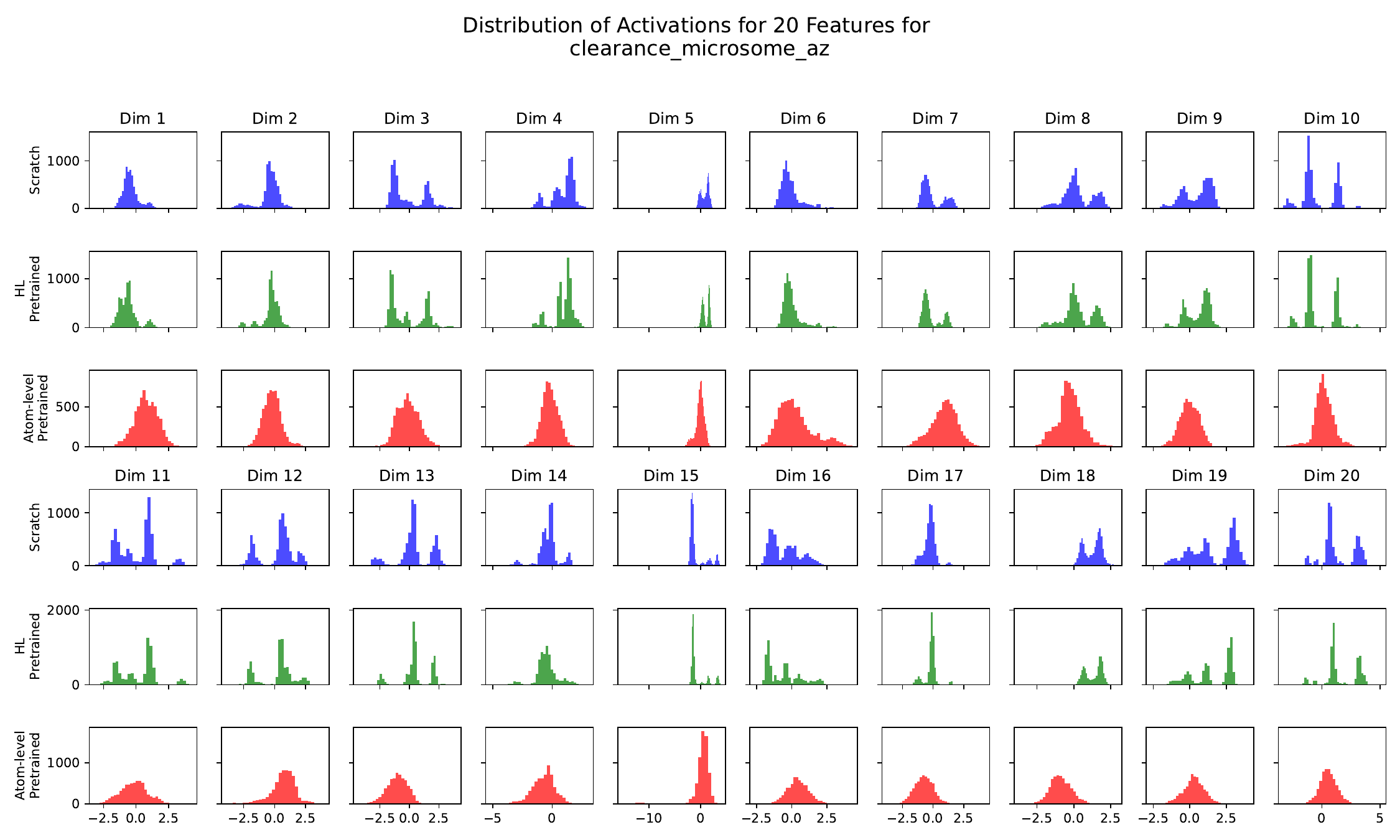}
    \caption{Distribution of first 20 features from the first layer of the Graphormer network for three different training approaches —scratch, HOMO-LUMO pretrained and atom-level pretrained— across test split of clearance microsome az dataset.}
    \label{fig:distributions_clearance_microsome_az}
\end{figure}

\begin{figure}
    \centering
    \includegraphics[width=\textwidth]{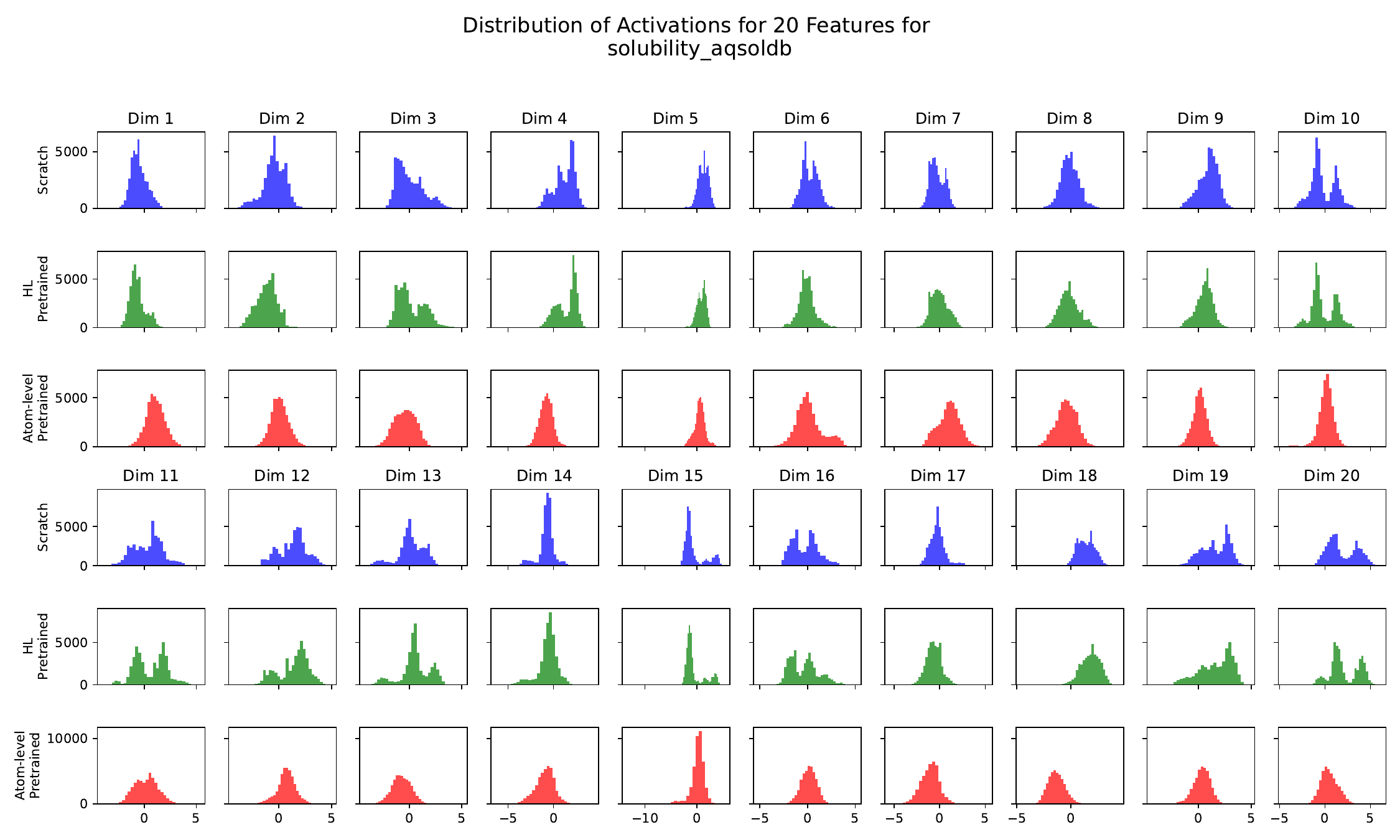}
    \caption{Distribution of first 20 features from the first layer of the Graphormer network for three different training approaches —scratch, HOMO-LUMO pretrained and atom-level pretrained— across test split of solubility aqsoldb dataset.}
    \label{fig:distributions_solubility_aqsoldb}
\end{figure}

\begin{figure}
    \centering
    \includegraphics[width=\textwidth]{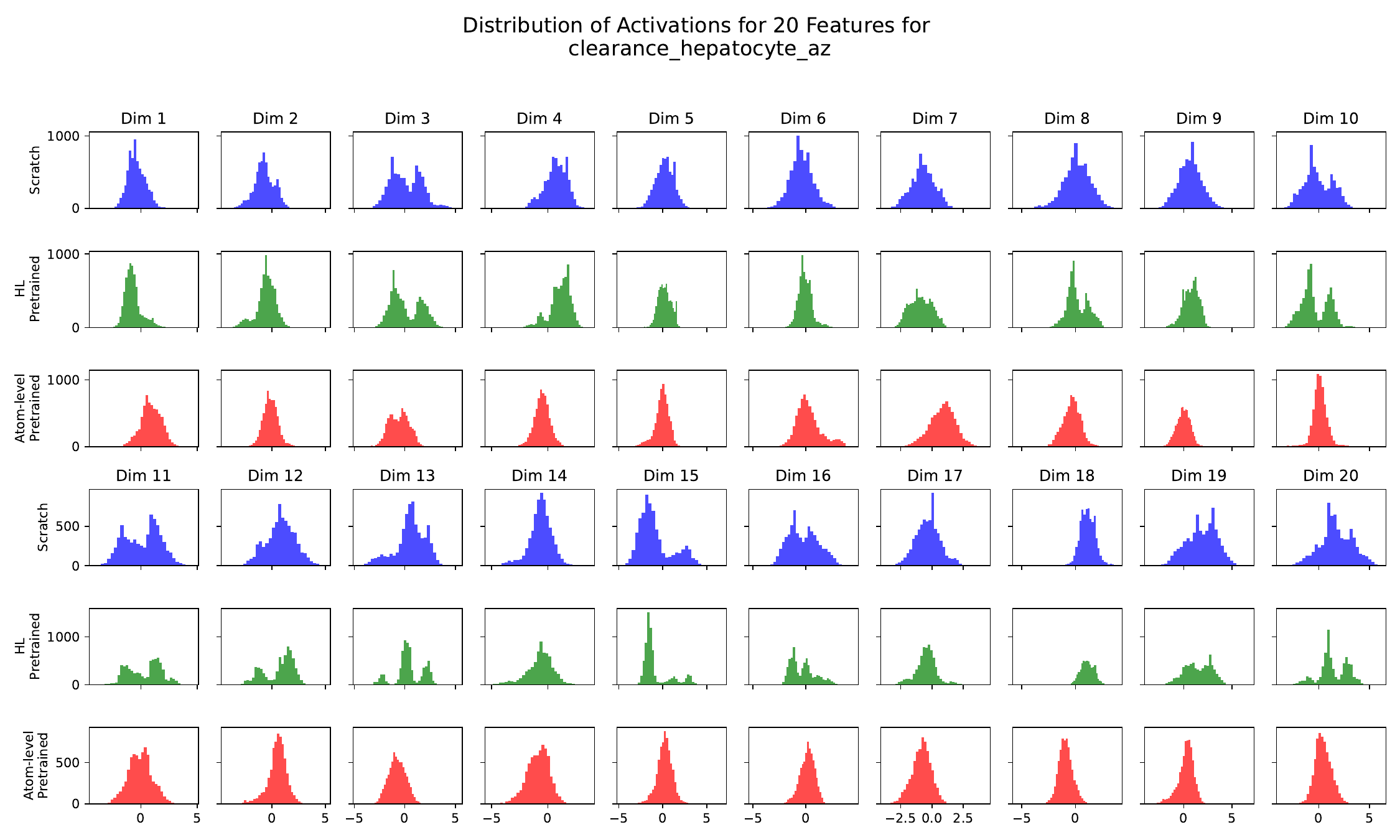}
    \caption{Distribution of first 20 features from the first layer of the Graphormer network for three different training approaches —scratch, HOMO-LUMO pretrained and atom-level pretrained— across test split of clearance hepatocyte az dataset.}
    \label{fig:distributions_clearance_hepatocyte_az}
\end{figure}

\begin{figure}
    \centering
    \includegraphics[width=\textwidth]{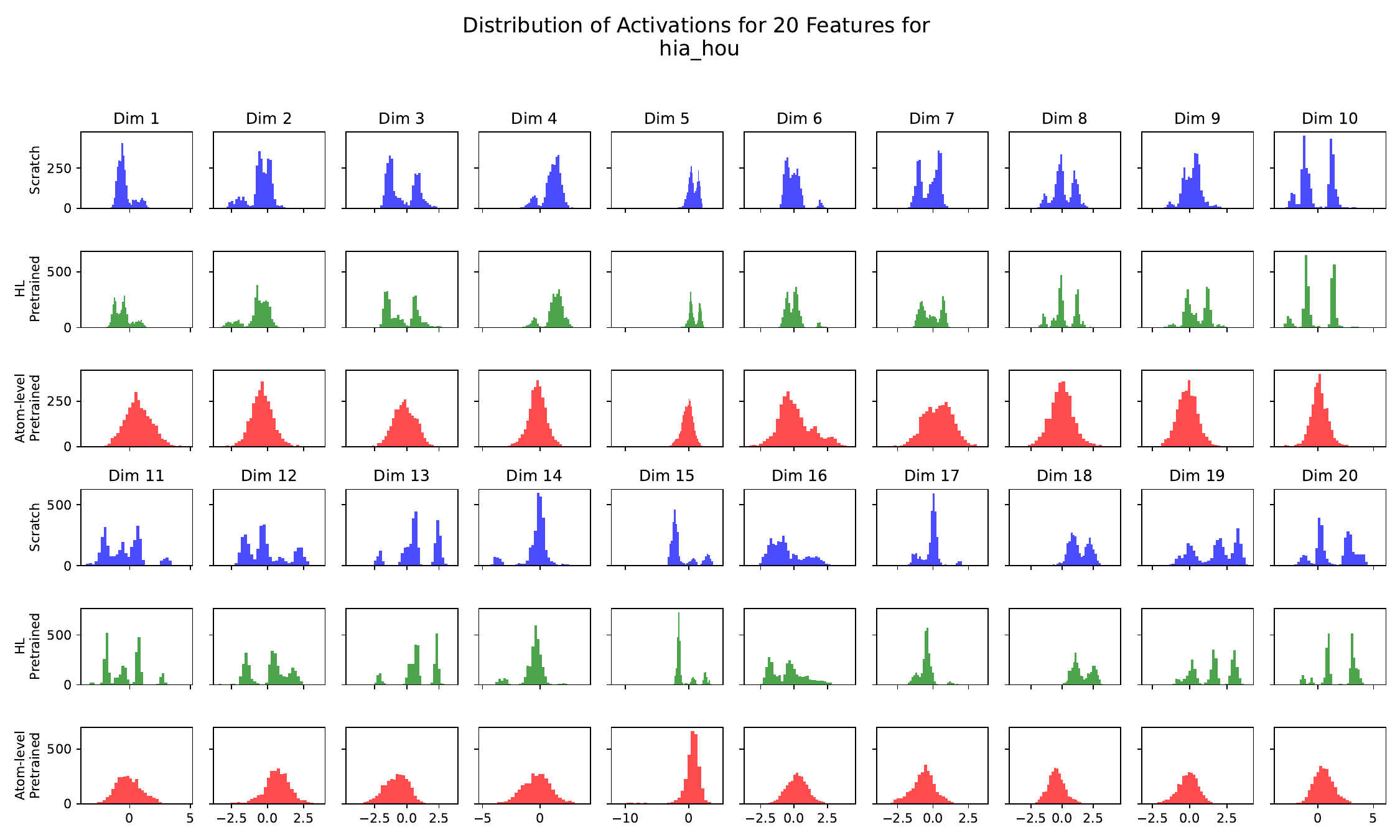}
    \caption{Distribution of first 20 features from the first layer of the Graphormer network for three different training approaches —scratch, HOMO-LUMO pretrained and atom-level pretrained— across test split of hia hou dataset.}
    \label{fig:distributions_hia_hou}
\end{figure}

\begin{figure}
    \centering
    \includegraphics[width=\textwidth]{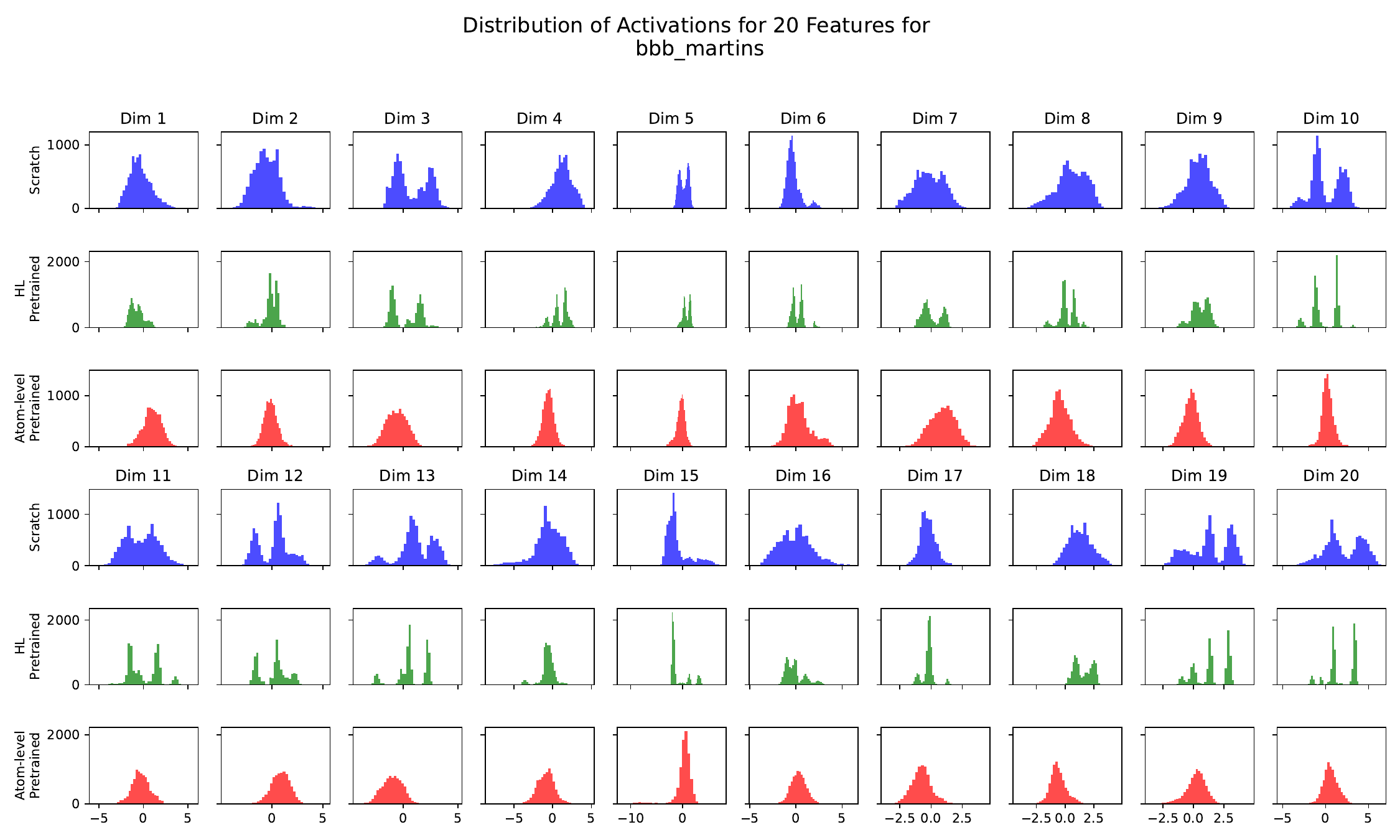}
    \caption{Distribution of first 20 features from the first layer of the Graphormer network for three different training approaches —scratch, HOMO-LUMO pretrained and atom-level pretrained— across test split of bbb martins dataset.}
    \label{fig:distributions_bbb_martins}
\end{figure}

\begin{figure}
    \centering
    \includegraphics[width=\textwidth]{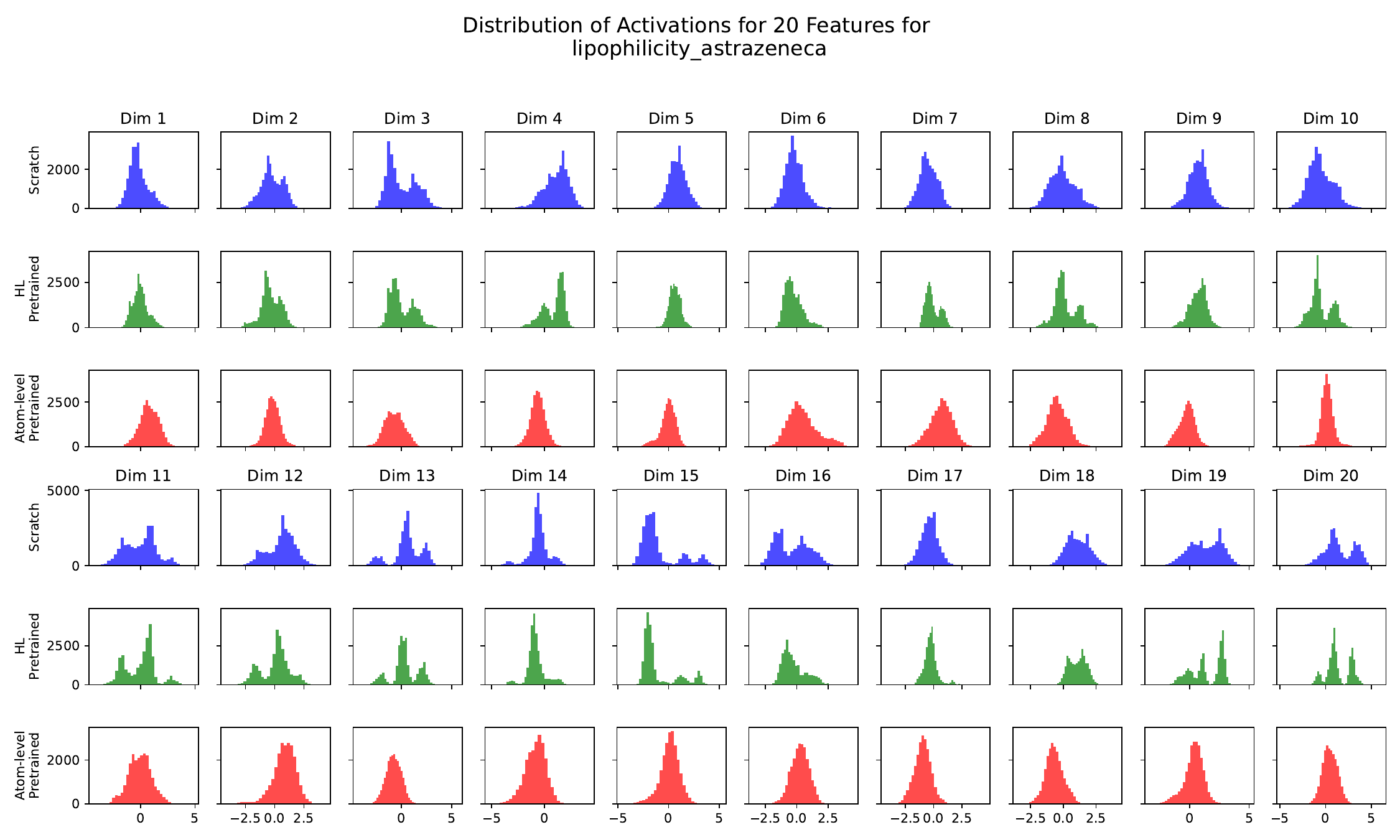}
    \caption{Distribution of first 20 features from the first layer of the Graphormer network for three different training approaches —scratch, HOMO-LUMO pretrained and atom-level pretrained— across test split of lipophilicity astrazeneca dataset.}
    \label{fig:distributions_lipophilicity_astrazeneca}
\end{figure}

\begin{figure}
    \centering
    \includegraphics[width=\textwidth]{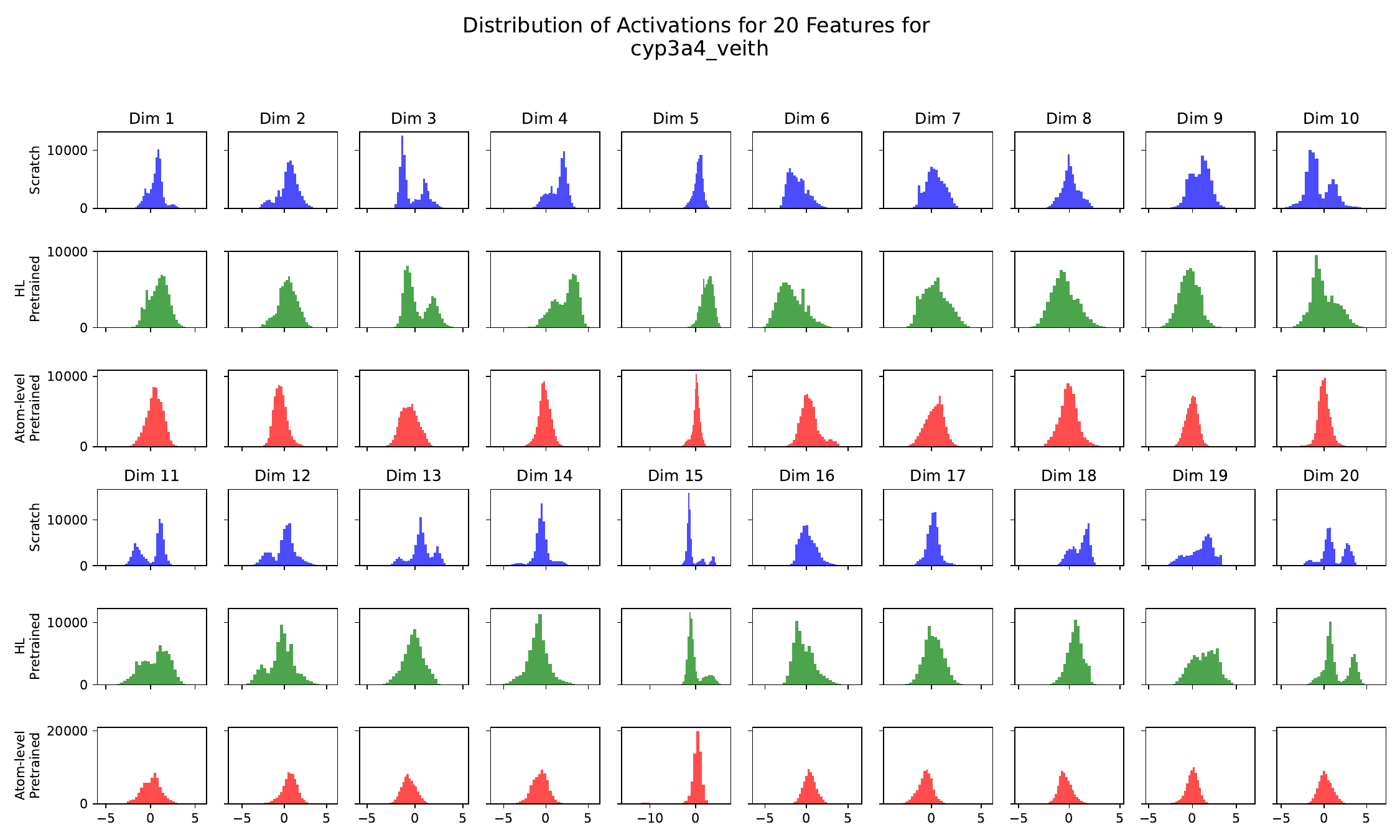}
    \caption{Distribution of first 20 features from the first layer of the Graphormer network for three different training approaches —scratch, HOMO-LUMO pretrained and atom-level pretrained— across test split of cyp3a4 veith dataset.}
    \label{fig:distributions_cyp3a4_veith}
\end{figure}

\begin{figure}
    \centering
    \includegraphics[width=\textwidth]{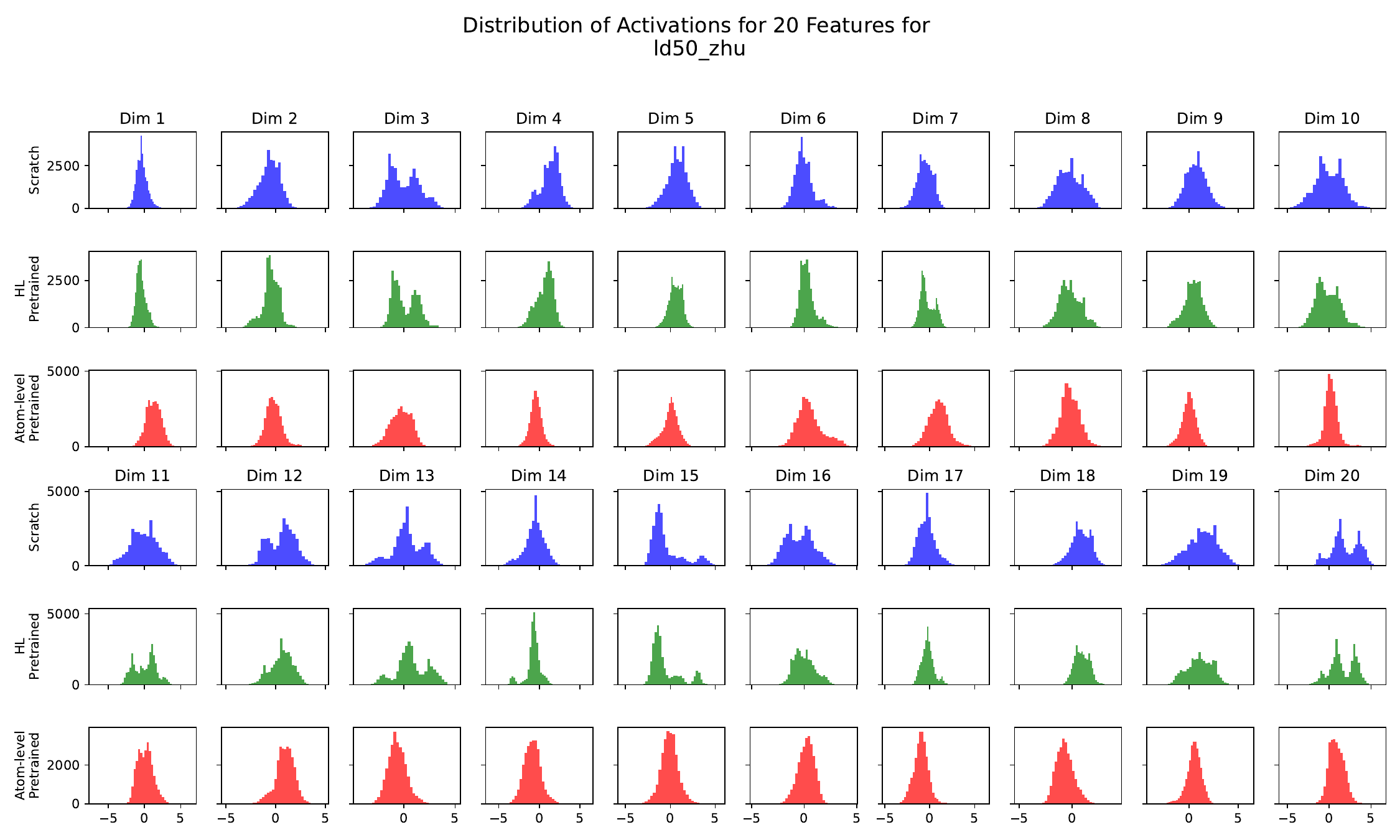}
    \caption{Distribution of first 20 features from the first layer of the Graphormer network for three different training approaches —scratch, HOMO-LUMO pretrained and atom-level pretrained— across test split of ld50 zhu dataset.}
    \label{fig:distributions_ld50_zhu}
\end{figure}

\begin{figure}
    \centering
    \includegraphics[width=\textwidth]{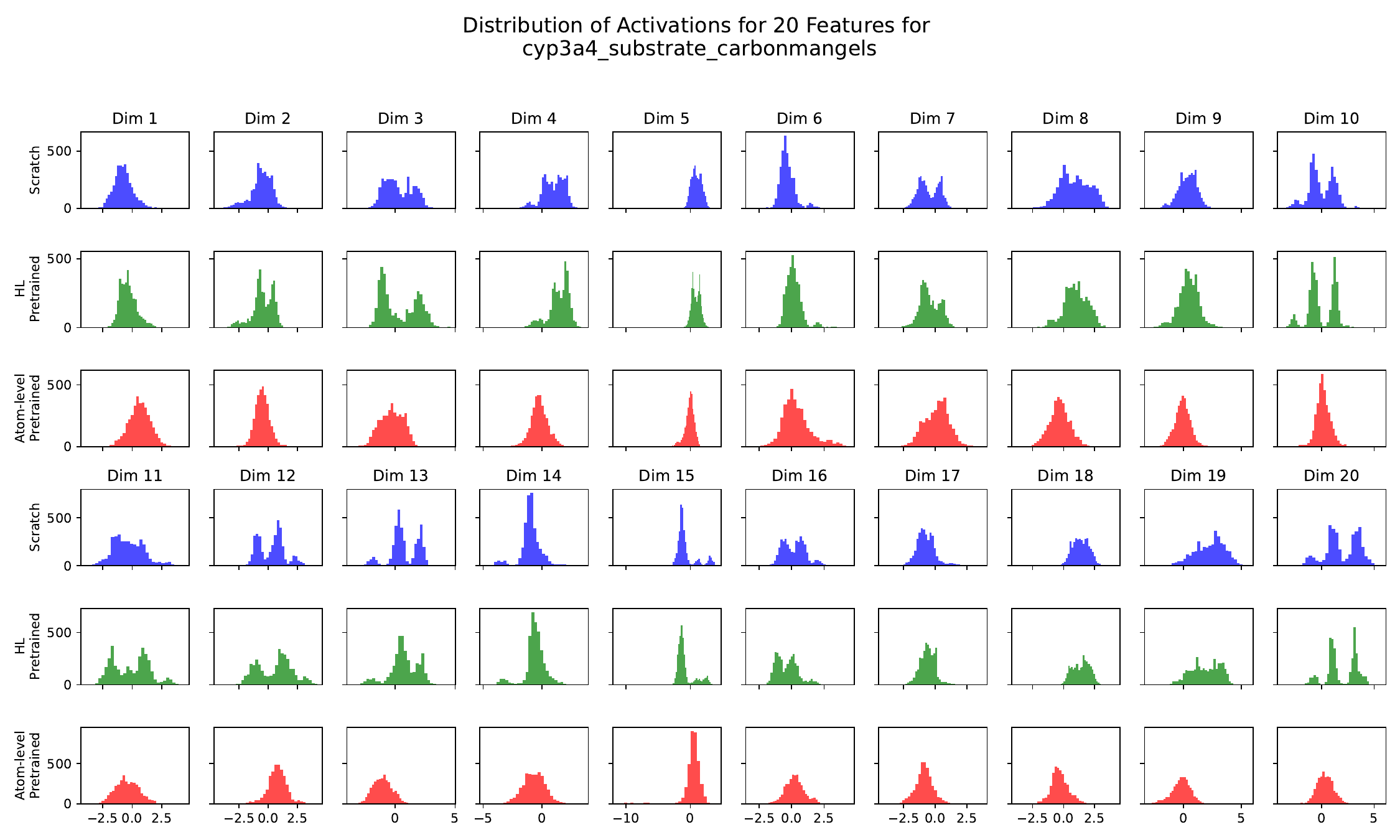}
    \caption{Distribution of first 20 features from the first layer of the Graphormer network for three different training approaches —scratch, HOMO-LUMO pretrained and atom-level pretrained— across test split of cyp3a4 substrate carbonmangels dataset.}
    \label{fig:distributions_cyp3a4_substrate_carbonmangels}
\end{figure}

\begin{figure}
    \centering
    \includegraphics[width=\textwidth]{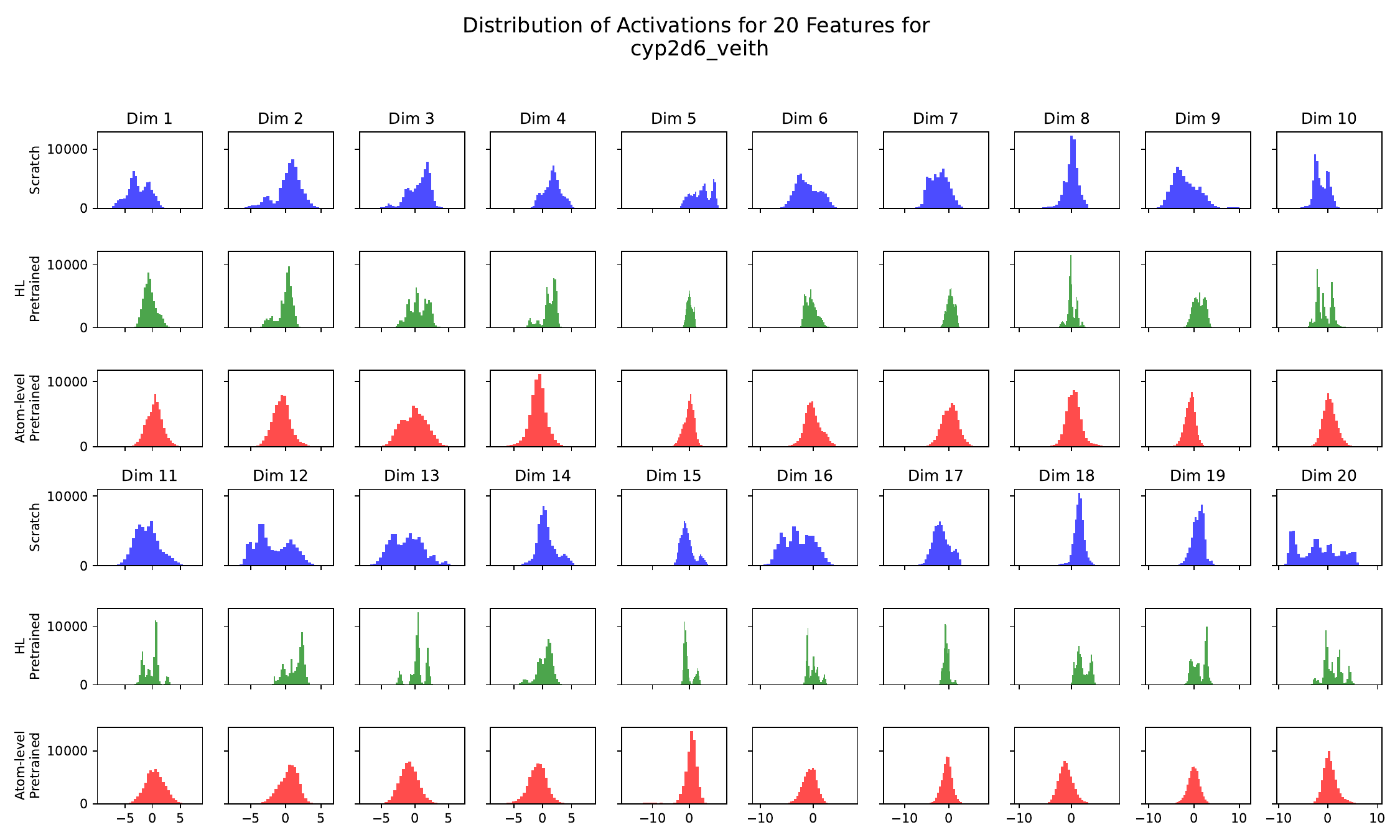}
    \caption{Distribution of first 20 features from the first layer of the Graphormer network for three different training approaches —scratch, HOMO-LUMO pretrained and atom-level pretrained— across test split of cyp2d6 veith dataset.}
    \label{fig:distributions_cyp2d6_veith}
\end{figure}

\begin{figure}
    \centering
    \includegraphics[width=\textwidth]{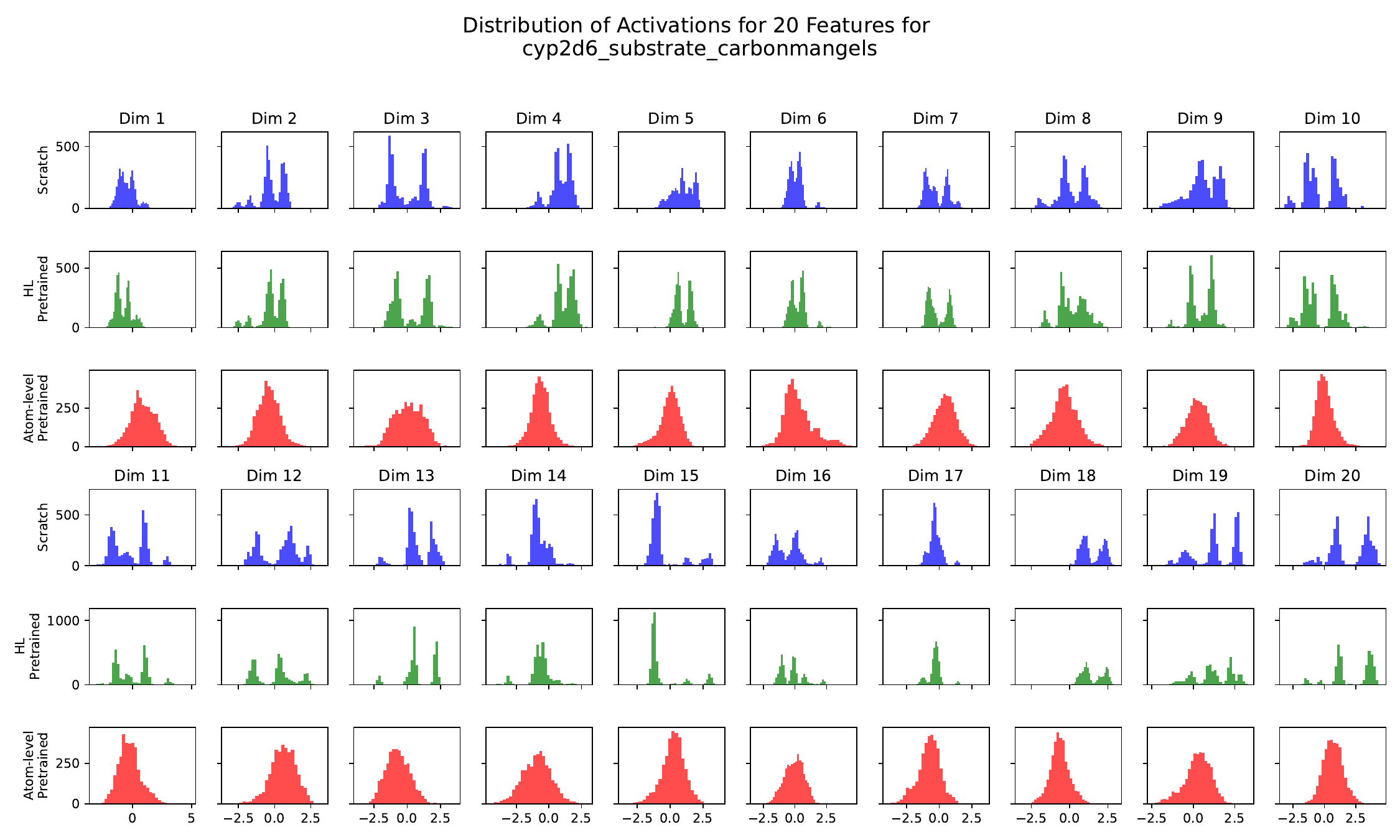}
    \caption{Distribution of first 20 features from the first layer of the Graphormer network for three different training approaches —scratch, HOMO-LUMO pretrained and atom-level pretrained— across test split of cyp2d6 substrate carbonmangels dataset.}
    \label{fig:distributions_cyp2d6_substrate_carbonmangels}
\end{figure}

\begin{figure}
    \centering
    \includegraphics[width=\textwidth]{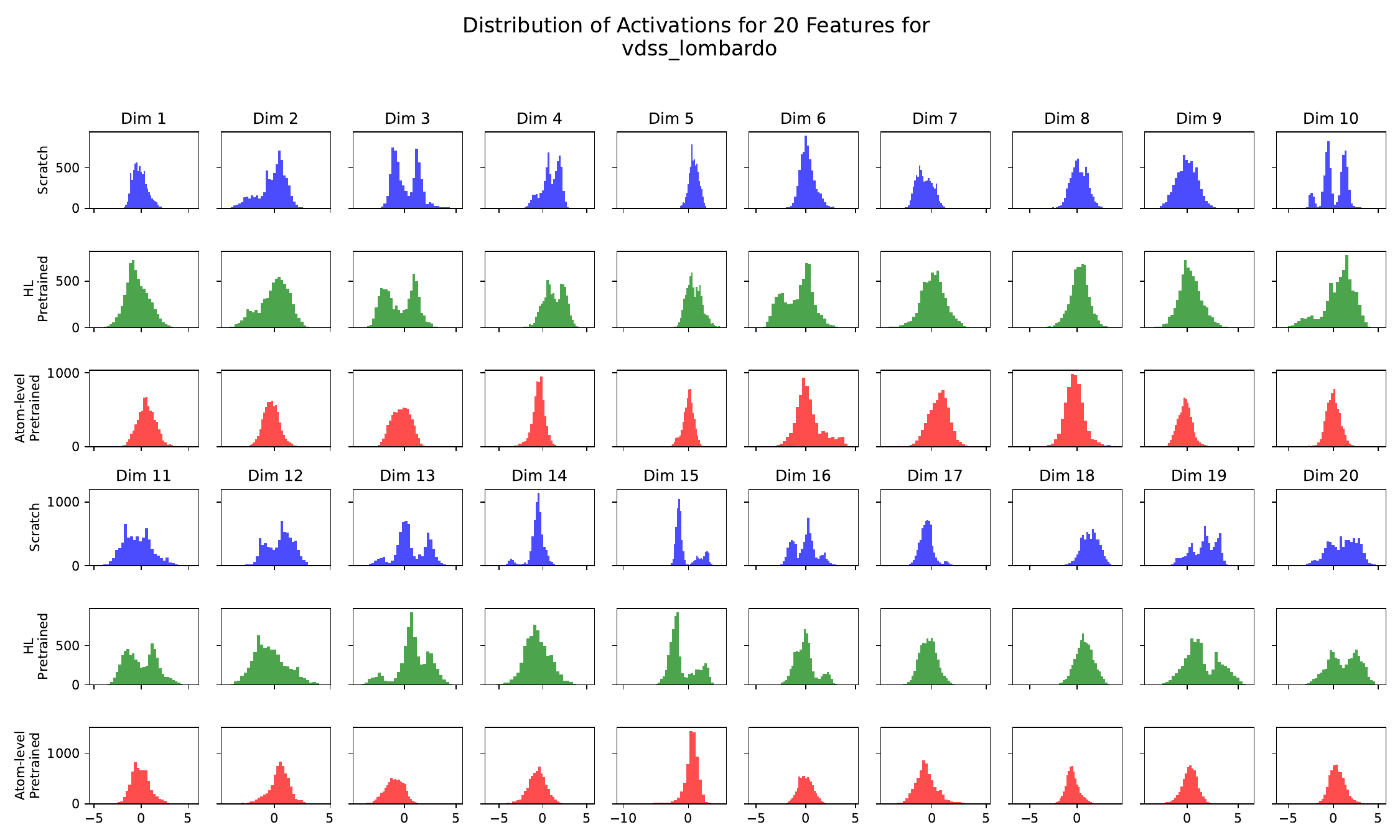}
    \caption{Distribution of first 20 features from the first layer of the Graphormer network for three different training approaches —scratch, HOMO-LUMO pretrained and atom-level pretrained— across test split of vdss lombardo dataset.}
    \label{fig:distributions_vdss_lombardo}
\end{figure}

\end{document}